%% file: ms.tex
\documentclass[10pt,twocolumn,letterpaper]{article}

\usepackage{cvpr}
\usepackage{times}
\usepackage{epsfig}
\usepackage{graphicx}
\usepackage{array}
\usepackage{rotating}
\usepackage{amsmath}
\usepackage{amsfonts}
\usepackage{hyperref}
\usepackage{color}
\usepackage{xfrac}
\usepackage{booktabs}
\usepackage{multirow}
\usepackage{float}
\usepackage{cuted}

\newcommand{\car}{\texttt{cars} }
\newcommand{\lane}{\texttt{lanes} }
\renewcommand{\vec}[1]{\mathbf{#1}}
\newcommand{\vz}{\vec{z}}
\newcommand{\vx}{\vec{x}}

\cvprfinalcopy

\begin{document}

\title{On the Road with 16 Neurons:\\Mental Imagery with Bio-inspired Deep Neural Networks}

\author{Alice Plebe\\
Dept. of Information Engineering and Computer Science\\
University of Trento, Italy\\
{\tt\small alice.plebe@unitn.it}
\and
Mauro Da Lio\\
Dept. of Industrial Engineeringe\\
University of Trento, Italy\\
{\tt\small mauro.dalio@unitn.it}
}

\maketitle


\begin{abstract}
\input{tex/abs}
\end{abstract}


\section{Introduction}
\label{s:into}
\input{tex/intro}

\section{Related Works}
\label{s:others}
\input{tex/others}

\section{The Neural Models}
\label{s:impl}
\input{tex/impl}

\section{Results}
\label{s:res}
\input{tex/res}

\section{Conclusions}
\label{s:end}
\input{tex/end}

\section{Acknowledgements}
\input{tex/ack}


\appendix{
\vspace*{70pt}
\section{Variational Inference}
\label{a:vae}
\input{tex/appx}
\section{Tables of Network Parameters}
\label{a:table}
\input{tex/table}
}

\bibliographystyle{ieee_fullname}

\input{ms.bbl}
\end{document}

%% file: tex/abs.tex
\noindent
This paper proposes a strategy for visual prediction in the context of autonomous driving. 
Humans, when not distracted or drunk, are still the best drivers you can currently find. For this reason we take inspiration from two theoretical ideas about the human mind and its neural organization.
The first idea concerns how the brain uses a hierarchical structure of neuron ensembles to extract abstract concepts from visual experience and code them into compact representations.
The second idea suggests that these neural perceptual representations are not neutral but functional to the prediction of the future state of affairs in the environment. Similarly, the prediction mechanism is not neutral but oriented to the current planning of a future action.
We identify within the deep learning framework two artificial counterparts of the aforementioned neurocognitive theories. We find a correspondence between the first theoretical idea and the architecture of convolutional autoencoders, while we translate the second theory into a training procedure that learns compact representations which are not neutral but oriented to driving tasks, from two distinct perspectives.
%
%
From a static perspective, we force groups of neural units in the compact representations to distinctly represent specific concepts crucial to the driving task. From a dynamic perspective, we encourage the compact representations to be predictive of how the current road scenario will change in the future.
We successfully learn compact representations that use as few as 16 neural units for each of the two basic driving concepts we consider: \car and \lane\hspace{-4pt}. We prove the efficiency of our proposed perceptual representations on the SYNTHIA dataset.
Our source code is available at \href{https://github.com/3lis/rnn_vae}{\tt{https://github.com/3lis/rnn\_vae}}.

%% file: tex/intro.tex
\begin{figure}
\begin{center}
\includegraphics[width=0.8760\columnwidth]{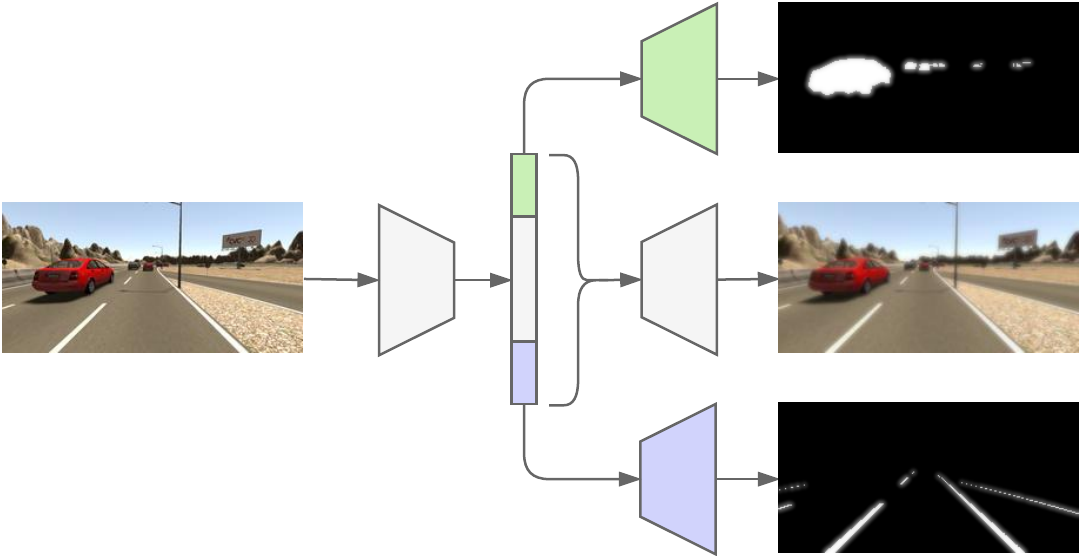}\\(a)\\
\vspace*{10pt}
\begin{tabular}{
c@{\hspace{5pt}}
c@{\hspace{5pt}}
c@{\hspace{5pt}}
c@{\hspace{5pt}}
}
\includegraphics[width=0.2160\columnwidth]{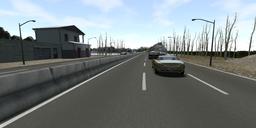} &
\includegraphics[width=0.2160\columnwidth]{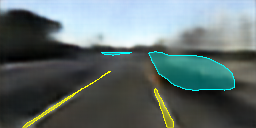} &
\includegraphics[width=0.2160\columnwidth]{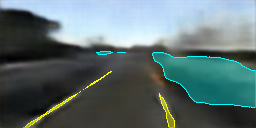} &
\includegraphics[width=0.2160\columnwidth]{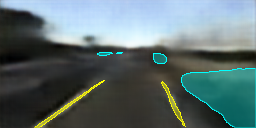}
\end{tabular}\\
\vspace*{3pt}
(b)\\
\vspace*{5pt}
\caption{\label{f:intro}\small (a) Our approach adopts a first model to learn a compact representation of the scenario, forcing groups of neurons to distinctly represent two basic concepts crucial to the driving task: \car and \lane. (b) The compact representations are used by a second network to predict future frames.}
\end{center}
\end{figure}

\noindent
Road traffic injuries are the leading cause of death for the age group between 5 and 29 years \cite{WHO:2018}. In the \textit{Global status report on road safety} of 2018, the World Health Organization reported that the number of road traffic deaths was 16 times larger than in war conflicts. This suggests that avoidance or mitigation of harm caused by motor vehicle accidents will be probably the most beneficial outcome one can expect from artificial intelligence and automation \cite{Fleetwood:2017}.  In fact, in the US only 2\% of vehicle crashes are due to technical failures. Among the major causes of accidents are driver's inattention, fast or reckless driving, illegal maneuvers, and tiredness \cite{Singh:2015}.

Self-driving cars will be immune to all the risky factors depending on human drivers. The development of fully autonomous vehicles has always be considered a coveted achievement for the modern society. The research on this field has a long history that dates back to the late 70s \cite{Dickmanns:1998}, but it became a reality -- at an unusually fast pace -- no longer than a decade ago \cite{InglePhute:2016}. While most of the components of a self-driving system (such as sensors) have improved at the typical rate of technological progress without any specific crucial innovations, the impressive advances have been mainly fueled by the emerging ``deep'' version of artificial neural networks \cite{HintonEtAl:2006,Schmidhuber:2015,LeCunEtAl:2015}.
Since their early beginnings, the greatest challenge for autonomous driving systems is the perception and understanding of the road environment, and this is precisely the most successful field of application of deep neural models \cite{LeCunEtAl:2010,SzegedyEtAl:2017,VanRullen:2017}. Therefore, deep neural models have quickly become the method of choice for driving scene perception \cite{CChenEtAl:2015,BojarskiEtAl:2017,GrigorescuEtAl:2019,WangEtAl:2019}. 
However, despite the impressive progress, perception remains the major obstacle towards fully autonomous vehicles. The core of this issue can be identify in the narrow conception of ``perception'' usually assumed in autonomous driving, which lacks the fundamental aspect of gathering knowledge about objects and events in the environment, to the point of being able making predictions for action \cite{Mesulam:1998,JacobJeannerod:2003}.

In this respect, it might be useful to reflect on how humans are able to drive. When not distracted, or asleep, or deliberately engaged in dangerous maneuvering, humans are excellent at driving, as at many other complex and highly specialized sensorimotor behaviors. How the brain realizes such sensorimotor behaviors is far from being fully understood, but there are few general neurocognitive theories trying to shed light on this. We deem it useful to borrow two theoretical ideas, in particular, to design the perception strategy of autonomous vehicles.

The first neurocognitive theory we take inspiration from concerns how sensorial information is coded into low-dimensional representations. These perceptual representations are able to capture aspects relevant to actions, and they allow their original content to be recreated in an approximated form during the phenomenon of mental imagery \cite{Kosslyn:1994,MoultonKosslyn:2009}.
One of the first piece of evidence of such representations was found in the work of Damasio \cite{Damasio:1989}, who identified neuron ensembles exhibiting a convergent structure, where neural signals are projected onto multiple cortical regions in a many-to-one fashion. Damasio later developed a broader theory \cite{MeyerDamasio:2009} identifying more sophisticated neural structures he called \textit{convergence-divergence zones} (CDZs). In this case, the very same neuron ensembles are able to perform both convergent and divergent projections, depending on the current action the brain is engaged with: the convergent flow is dominant during perceptual recognition, while the divergent flow occurs during mental imagery.
CDZs were recognized as a crucial component in the formation of concepts in the brain \cite{OlierEtAl:2017}. Therefore, we believe it useful to design a model with a similar hierarchical architecture to learn the abstract concepts relevant to the driving context.

The second theoretical idea concerns the nature of the neural representations in the brain. In most cases, neural representations are not neutral but functional to the prediction of the state of affairs in the future environment. Actually, the ability to predict appears to be the main goal of intelligence \cite{Jeannerod:2001,Hesslow:2012}.
There is evidence for the existence of various circuits in the brain that provide prediction from perceptual representations. In particular, two forms of prediction -- procedural and declarative -- are typically acknowledged in different brain structures \cite{Downing:2009}.
However, one of the most popular theories in the field interprets the mental mechanism of prediction in mathematical terms \cite{Friston:2010,FristonEtAl:2017}. This theory, called \textit{predictive brain}, explains the behavior of the brain as minimization of the free-energy, a quantity that can be expressed in mathematical form. We will show how this formulation can actually be adopted as loss function to train our model.

The aim of our work is to learn conceptual representations of the driving scenario from visual information. Our intention is to learn compact and informative representations that can be useful for a variety of downstream driving tasks.
We propose a cognitive-inspired approach that enforces the representations to be not neutral but oriented to the driving tasks, under two distinct perspectives.
From a static perspective, we force groups of neural units in the compact representation to distinctly represent specific concepts which are crucial in the driving task. Specifically, we use as few as 16 neurons for each of the two basic concepts: \car and \lane\hspace{-2pt}.
From a dynamic perspective, we encourage the compact representations to be predictive of how the current road scene would change in the future.

We achieve the conceptual representations by implementing an artificial neural model that is in line with the two aforementioned neurocognitive theories.
The term ``neural'' in artificial neural models by no means implies a faithful replication of the computations performed by biological neurons. On the contrary, the mathematics of deep learning bears little resemblance to the way brain works \cite{Rolls:2016,Conway:2018}.
However, we identify two methods within the framework of artificial neural networks (ANNs) that appear, at least in part, rough algorithmic counterparts of the neurocognitive theories described above. Specifically, the CDZs may find a correspondence in the idea of convolutional autoencoders \cite{TschannenEtAl:2018}, while the predictive brain theory resonates with the adoption of Bayesian variational inference in combination with autoencoders \cite{KingmaWelling:2014,RezendeEtAl:2014}.

This work is part of the H2020 Dreams4Cars\footnote{\href{http://dreams4cars.eu}{www.dreams4cars.eu}} project, aimed at developing an artificial driving agent inspired by the neurocognition of human driving \cite{PlebeEtAl:2019b}.
In the following section we discuss the most significant related works. In \S\ref{s:impl} we describe the implementation of 4 different neural models that successfully learn informative and compact representations. Lastly, Section \S\ref{s:res} presents the results of our models on the SYNTHIA dataset.

%% file: tex/others.tex
\noindent
It is not uncommon for works adopting ANNs for perception in autonomous vehicles to declare virtues of a neurocognitive inspiration \cite{PasquierOentaryo:2008,ChenEtAl:2017,ZhangEtAl:2019}. Among these are important claims, but they do not transfer the specific brain mechanisms into algorithms. To the best of our knowledge, the two neurocognitive principles embraced by this work -- Damasio's CDZs and Friston's predictive brain -- have not been proposed in any work on perception for autonomous driving. In addition, the striking similarity between the formulation of brain predictivity given by Friston and the variational autoencoder algorithm seems to remain unnoticed, with few exceptions \cite{OfnerStober:2018}.

The idea of autoencoder has been at the heart of the ``deep'' turn of ANNs \cite{HintonSalakhutdinov:2006,KrizhevskyHinton:2011,HintonEtAl:2011}, and their variational version has rapidly gained attention \cite{CZhangEtAl:2019}. However, it has not yet been widely adopted for autonomous vehicle perception. One of the most popular strategies, instead, is the \textit{end-to-end} approach, where images from a front-facing camera are fed into a stack of convolutions followed by feedforward layers, generating low-level commands. The first attempt in this direction dates before the rise of deep learning \cite{MullerEtAl:2006}, and it has been the groundwork for the later NVIDIA's PilotNet \cite{BojarskiEtAl:2016,BojarskiEtAl:2017}.  One of the most serious drawbacks of end-to-end systems based on static frame processing is the erratic variation of steering wheel angle within short time periods. A potential solution is to provide temporal context in the models, combining convolutions with recurrent networks \cite{EraqiEtAl:2017}.

But the most appealing feature of the end-to-end strategy -- to dispense with internal representations -- is also the major source of its troubles. Learning the entire range of road scenarios from steering supervision alone, with all possible appearances of objects relevant to the drive, is not achievable in practical settings.  For this reason several more recent proposals suggest inclusion of intermediate representations, such as the so-called \textit{mid-to-mid} strategy \cite{BansalEtAl:2018} used in ChauffeurNet, Waimo's autonomous driving system. ChauffeurNet is essentially made of a convolutional network which consumes the input data to generate an intermediate representation with the format of top-down view of the surrounding area and salient objects. In addition, ChauffeurNet has several higher-level networks which iteratively predict information useful for driving.
In \cite{DWangEtAl:2019} the main aim is to overcome the object agnosticism of the end-to-end approach, the Authors propose a \textit{object-centric} deep learning system for autonomous vehicles. In their proposal there is one convolutional neural module taking an RGB image and producing an intermediate representation. Then, the downstream networks are diversified depending on a taxonomy of objects-related structures in the intermediate representation, which are lastly converted into discrete driving actions.
An internal representation is used also in the system by Valeo Vision \cite{SistuEtAl:2019}, constructed using a standard ResNet50 model \cite{HeEtAl:2016} with the top fully-connected layers removed. The feature representation is shared across a multitude of tasks relevant to visual perception in automated driving such as object detection, semantic segmentation, depth estimation. All the downstream tasks are realized using the top parts of standard models such as YOLO \cite{RedmonFarhadi:2018} for object detection or FCN8 \cite{LongEtAl:2015} for semantic segmentation.

None of the works reviewed so far builds the internal representations through the idea of autoencoder. We found just two notable exceptions in the field of perception for autonomous driving. The first one is by \textit{comma.ai} \cite{SantanaHotz:2016}, in their model the latent representation has dimension 2048 neurons, and it is obtained with a variational autoencoder that restores the input images of $160\times80$ pixels with 4 deconvolutional layers. Once trained, the latent representations are used for predicting successor frames in time with a recurrent neural network. The second exception is a work by Toyota in collaboration with MIT \cite{AminiEtAl:2019}, using a variational autoencoder of dimension 25 neurons. This entire internal representation is decoded to restore the input image of size $200\times66$ as in a standard autoencoder. In addition, one neuron of the representation is interpreted as steering angle, therefore an end-to-end supervision for this neuron is mixed in the total training loss.

There are similarities between these last two approaches and the one we present, but also fundamental differences. The latent compact representation of Amini et al. does not take into account the crucial time dimension of the perceptual driving scenario. On the other hand, Santana and Hotz include their internal representation in a recursive network for prediction, but time dependency is not exploited when learning the compact representation. Moreover, the \textit{comma.ai}'s model is agnostic about the meaning of the neurons composing the latent representation, while Amini et al. assign meaning to just the single neuron coding steering angles. One key strategy in our model is to encourage the assignment of conceptual meaning to segregate groups of neurons in the latent representation. In contexts different from autonomous vehicles the idea is not new. For example, \cite{KulkarniEtAl:2015} in human heads generation \cite{KulkarniEtAl:2015} proposed a latent space with separate representations for viewpoints, lighting conditions, and shape variations. Also in \cite{ZhaoEtAl:2016} the latent vector is partitioned in semantic content and geometric coding.

%% file: tex/impl.tex
\noindent
During the development of this work, we experimented a number of different architectures, all sharing the common feature of an hierarchical arrangement similar to the CDZs in the brain, according to the strategy described in the Introduction.
The first group of neural models was developed with the aim of generating a latent representation of the driving environment. The challenge here is to ensure the latent space is informative enough to represent the wide variety of driving scenarios, but keeping at the same time a low dimensionality so that it can still be inspected and explicable.

Here we present:
\begin{itemize}
\item three different neural network (\textit{Net1, Net2, Net3}) with \textit{encoder--decoder} architectures, adopting increasingly sophisticated approaches to learn compressed and disentangled latent representations,
\item a fourth neural network (\textit{Net4}) performing prediction in time of future driving scenarios as a rudimentary form of mental imagery, working exclusively within the latent representation created by the three previous models.
\end{itemize}

\subsection{Net1: Variational Autoencoder}
\label{ss:vae}
\noindent
When talking about representation learning, the first architecture that comes to mind is the \textit{autoencoder}.
This is the simplest model of the family, composed of two sub-networks:
\begin{eqnarray}
g_\Phi&:\mathcal{X}\rightarrow\mathcal{Z},\label{e:encoder}\\
f_\Theta&:\mathcal{Z}\rightarrow\mathcal{X}.\label{e:decoder}
\end{eqnarray}
The first sub-network is called \textit{encoder} and computes the compact representations $\vz\in\mathcal{Z}$ of a high dimensional input $\vx\in\mathcal{X}$. This network is determined by its set of parameters $\Phi$.
The second sub-network is the \textit{decoder}, often called the \textit{generative} network, which reconstructs high--dimensional data $\vx\in\mathcal{X}$ taking as input low dimensional compact representations $\vz\in\mathcal{Z}$. The network is fully determined by the set of parameters $\Theta$.
When training the autoencoder, the parameters $\Theta$ and $\Phi$ are learned by minimizing the error between input samples $\vx_i$ and the outputs $f(g(\vx_i))$.

The next big improvement in the field of representation learning is the \textit{variational autoencoder}, we refer to \ref{a:vae} for a detailed mathematical definition.
The variational autoencoder is able to learn a more ordered representation with respect to the standard autoencoder. However, there is a lot space for improvements, especially in our case where we want to focus only on learning representations of driving scenarios. Therefore, we implemented the variational autoencoder model mainly with the intent to be used as a baseline for further improvements and comparison.

Table \ref{t:vae} shows the numbers of layers and the parameters adopted in the final version of the variational autoencoder (\textit{Net1}).
The input of the network is a single RGB image of $256\times256$ pixels, the encoder is composed of a stack of 4 convolutions and 2 fully-connected layers, converging to a latent space of 128 neurons. The decoder has a symmetric structure with respect to the encoder, mapping the 128 neurons back to an image of $256\times256$. The network is trained to optimize the loss function in equation \eqref{e:loss} in a totally unsupervised way.
%
%
%
%
%
%
\begin{figure}
\begin{center}
\includegraphics[height=200pt]{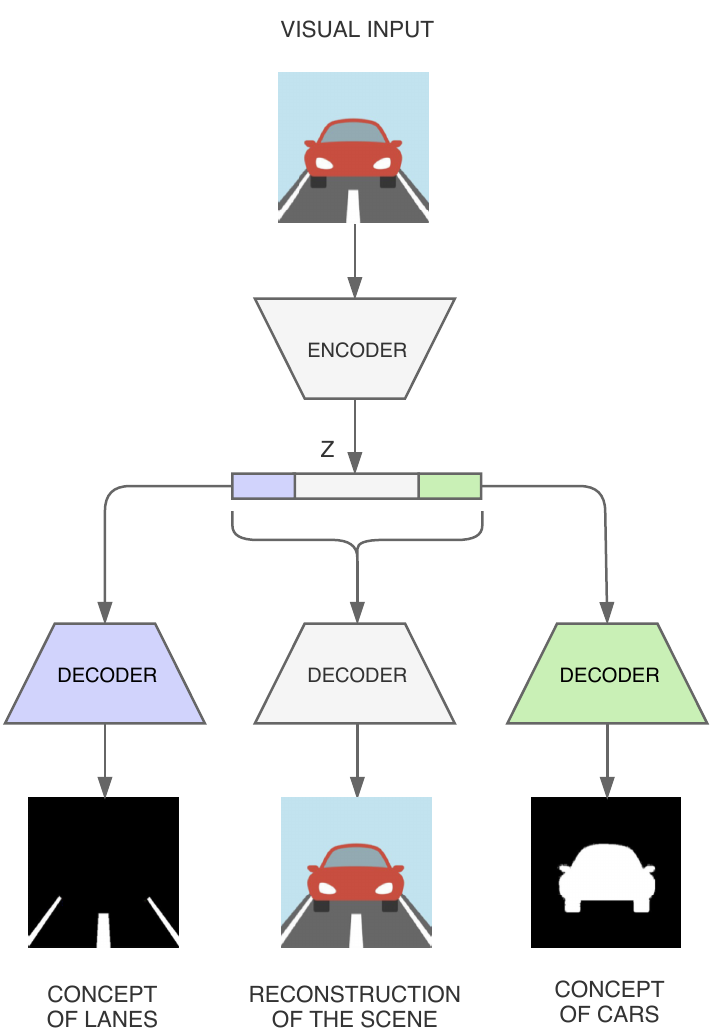}
\end{center}
\vspace{-5pt}
\caption{\label{f:mvae}\small Architecture of our topological autoencoder (\textit{Net2}), where the green color denotes the \car concept, violet the \lane concept.}
\end{figure}
%

\subsection{Net2: Topological Autoencoder}
\label{ss:mvae}
\noindent
As discussed in the Introduction, our brain naturally projects sensorial information -- especially visual -- into CDZs representations, including all those that constitute the \textit{conceptual space}, where neural activations represent the nature of entities present in the environment that produced the perceptual stimuli. Hence, we can take inspiration from this theory and use the CDZs hierarchical architecture as a ``blueprint'' to design a more sophisticated neural network, able to learn representations that are not only in terms of visual features, but also in terms of \textit{concepts}.
In the driving context the entire road scenario is informative, however, from a conceptual point of view it is not immediately necessary to infer categories for every entity present in a scene. It is useful to project in conceptual space only the entities mostly relevant to the driving task. Therefore, for simplicity, in this model we choose to consider the two main concepts of \car and \lane.

Fig. \ref{f:mvae} depicts the architecture of this topological autoencoder (\textit{Net2}), composed of one shared encoder and three independent decoders. The choice of parameters is similar to the architecture of our variational autoencoder (\textit{Net1}), Table \ref{t:mvae} shows the parameters of the final model considered. The encoder and each of the 3 decoders maintain the same structure as before, even the overall number of neurons in the latent space is the same, it is their inner organization that is strongly improved.
As shown in Fig. \ref{f:mvae}, the grey decoder is the one reconstructing in the visual space -- just like the decoder of \textit{Net1} -- mapping all the 128 neurons of the latent vector $\vz$ into an RGB image. This decoder learns to reconstruct the input image therefore is trained in a totally unsupervised way.
The decoder colored in green, instead, takes only a vector $\vz_\mathrm{C}$ of 16 neurons of the latent space and produces a matrix $\vx_\mathrm{C}$ of $256\times256$ probability values. The latent vector of 16 neurons is trained to represent the \car concept, and the output matrix can be interpreted as a semantic segmentation of the input image, where values indicates the probability of presence of \car entities. Similarly, the violet decoder maps only the vector $\vz_\mathrm{L}$ of 16 neurons representing the \lane concepts into a probability matrix $\vx_\mathrm{L}$ for \lane entities. These two decoders require a supervised learning: their output is converted into binary images, by applying a threshold, and trained to minimize the reconstruction error with semantic segmentation of the input images.

Note that the idea of partitioning the entire latent vector into meaningful components is not new, as mentioned in \S\ref{s:others}.
But our approach is different: while we keep the two segments $\vz_\mathrm{C}$ and $\vz_\mathrm{L}$ disjointed, it is the entire $\vz$ that learns representations in visual space. This way, we adhere entirely to the CDZ theoretical idea, and explicitly encourage the network to pay attention to the \car and \lane entities in the environment.
We would like to highlight another advantage of our approach in partitioning the latent space, concerning the crucial issue of lack of transparency in deep neural network. In most models no information is available about what exactly make them arrive at their predictions \cite{SamekEtAl:2017,PlebeEtAl:2019a}. By assigning meaning to components of the inner representation, the issue is mitigated.

To give a mathematical description, the overall model is composed of 4 sub-networks:
\begin{eqnarray}
g_\Phi&:&\mathcal{X}\rightarrow\mathcal{Z},
\nonumber\\
f_{\Theta_\mathrm{V}}&:&\mathcal{Z}\rightarrow\mathcal{X},
\nonumber\\
f_{\Theta_\mathrm{C}}&:&\mathcal{Z}_\mathrm{C}\rightarrow\mathcal{X}_\mathrm{C},
\nonumber\\
f_{\Theta_\mathrm{L}}&:&\mathcal{Z}_\mathrm{L}\rightarrow\mathcal{X}_\mathrm{L},
\nonumber
\end{eqnarray}
where the subscript $V$ denotes the visual space, and the subscripts $C$ and $L$ refer to the \car and \lane concepts respectively. For each vector $\vz$ in the latent space:
\begin{align}
\vz\in\mathcal{Z}&=\left[\vz_{\mathrm{C}}, \widetilde{\vz}, \vz_{\mathrm{L}}\right],
\label{e:z}\\
\mathcal{Z}&=\mathbb{R}^{N_\mathrm{V}},
\nonumber\\
\mathcal{Z}_\mathrm{C}&=\mathbb{R}^{N_\mathrm{C}},
\nonumber\\
\mathcal{Z}_\mathrm{L}&=\mathbb{R}^{N_\mathrm{L}},
\nonumber
\end{align}
In the first expression, $\vz_\mathrm{C}$ and $\vz_\mathrm{L}$ are the two segments inside the latent vector $\vz$ representing the \car and \lane concepts, respectively. The segment in between, $\widetilde{\vz}$, encodes the remaining generic visual features, and the entire latent vector $\vz$ is a representation in the visual space. In the final version of the model, we choose to have $N_\mathrm{V}=128$ and $N_\mathrm{C}=N_\mathrm{L}=16$.
%

By calling $\Theta=\left[\Theta_\mathrm{V},\Theta_\mathrm{C},\Theta_\mathrm{L}\right]$ the vector of all parameters in the three decoders, the loss functions of the model is derived from the basic equation \eqref{e:loss}. At each batch iteration $b$, a random batch $\mathcal{B}\subset\mathcal{D}$ is presented, and the following loss is computed:
\begin{equation}
\mathcal{L}(\Theta,\Phi|\mathcal{B})=E_\mathrm{K}+E_\mathrm{V}+E_\mathrm{C}+E_\mathrm{L}
\label{e:mloss}
\end{equation}
where
\begin{align}
E_\mathrm{K}=&\left(1-(1-k_0)\kappa^b\right)\sum_{\vx}^\mathcal{B}\Delta_{\mathrm{KL}}\big(q_{\Phi}(\vz|\vx)\|p_{\Theta_\mathrm{V}}(\vz)\big),
\label{e:mlossk}\\
E_\mathrm{V}=&-\lambda_\mathrm{V}\sum_{\vx}^\mathcal{B}\mathbb{E}_{\vz\sim{q_{\Phi}(\vz|\vx)}}\left[\log{p_{\Theta_\mathrm{V}}(\vx|\vz)}\right],
\label{e:mlossv}\\
E_\mathrm{C}=&-\lambda_\mathrm{C}\sum_{\vx}^\mathcal{B}\mathbb{E}_{\vz_\mathrm{C}\sim\Pi_\mathrm{C}({q_{\Phi}(\vz|\vx)})}\left[\log{\widetilde{p}_{\Theta_\mathrm{C}}}(\vx_\mathrm{C}|\vz_\mathrm{C})\right],
\label{e:mlossc}\\
E_\mathrm{L}=&-\lambda_\mathrm{L}\sum_{\vx}^\mathcal{B}\mathbb{E}_{\vz_\mathrm{L}\sim\Pi_\mathrm{L}({q_{\Phi}(\vz|\vx)})}\left[\log{\widetilde{p}_{\Theta_\mathrm{L}}}(\vx_\mathrm{L}|\vz_\mathrm{L})\right].
\label{e:mlossl}
\end{align}
Few observations are due for the differences between this loss function and the basic one \eqref{e:loss}. First of all, there is a delay in including the contribution of the Kullback-Leibler divergence in the term $E_\mathrm{K}$, because initially the encoder is unlikely to provide any meaningful probability distribution $q_{\Phi}(\vz|\vx)$. There is a cost factor for the KL component, set initially at a small value $k_0$ and gradually increased up to $1.0$, with time constant $\kappa$. This strategy is called \textit{KL annealing} and was first introduced in the context of variational autoencoders for language modeling \cite{BowmanEtAl:2015}.

The remaining terms $E_\mathrm{V},E_\mathrm{C},E_\mathrm{L}$ are errors in the reconstruction of driving scenario and conceptual entities, and their relative contributions are weighted by the parameters $\lambda_\mathrm{V},\lambda_\mathrm{C},\lambda_\mathrm{L}$. The purpose of these parameters is mainly to normalize the range of the errors, which is quite different from visual to conceptual spaces. For this reason, typically $\lambda_\mathrm{V}\ne\lambda_\mathrm{C}=\lambda_\mathrm{L}$.
The term $E_\mathrm{V}$ computes the error in visual space, using the entire latent vector $\vz$, and corresponds precisely to the second component in the basic loss \eqref{e:loss}. The last two terms $E_\mathrm{C}$ and $E_\mathrm{L}$ compute the error in the conceptual space and are slightly different. Only the relevant portion of the latent vector $\vz$ is considered, as indicated by the projection operators $\Pi_\mathrm{C},\Pi_\mathrm{L}$. In addition, a variant of the standard cross entropy is used, indicated with the symbols $\widetilde{p}_{\Theta_\mathrm{C}}$ and $\widetilde{p}_{\Theta_\mathrm{L}}$, in order to account for the large unbalance between the number of pixels belonging to a concept and all the other pixels, which is typical in ordinary driving scenes. Following the method first introduced in the context of medical image processing \cite{SudreEtAl:2017}, we compensate this asymmetry by weighing the contribution of true and false pixels with $P$, the ratio of true pixels over all the pixels in the dataset, computed as follows:
\begin{equation}
P=\left(\frac{1}{NM}\sum^N_i\sum^M_jy_{i,j}\right)^\frac{1}{s},
\label{e:P}
\end{equation}
where $N$ is the number of pixels in an image, $M$ is the number of images in the training dataset.  The parameter $s$ is used to smooth the effect of weighting by the probability of ground truth, a value evaluated empirically as valid is $4$.

\begin{figure*}[t]
\begin{center}
\includegraphics[width=1.481\columnwidth]{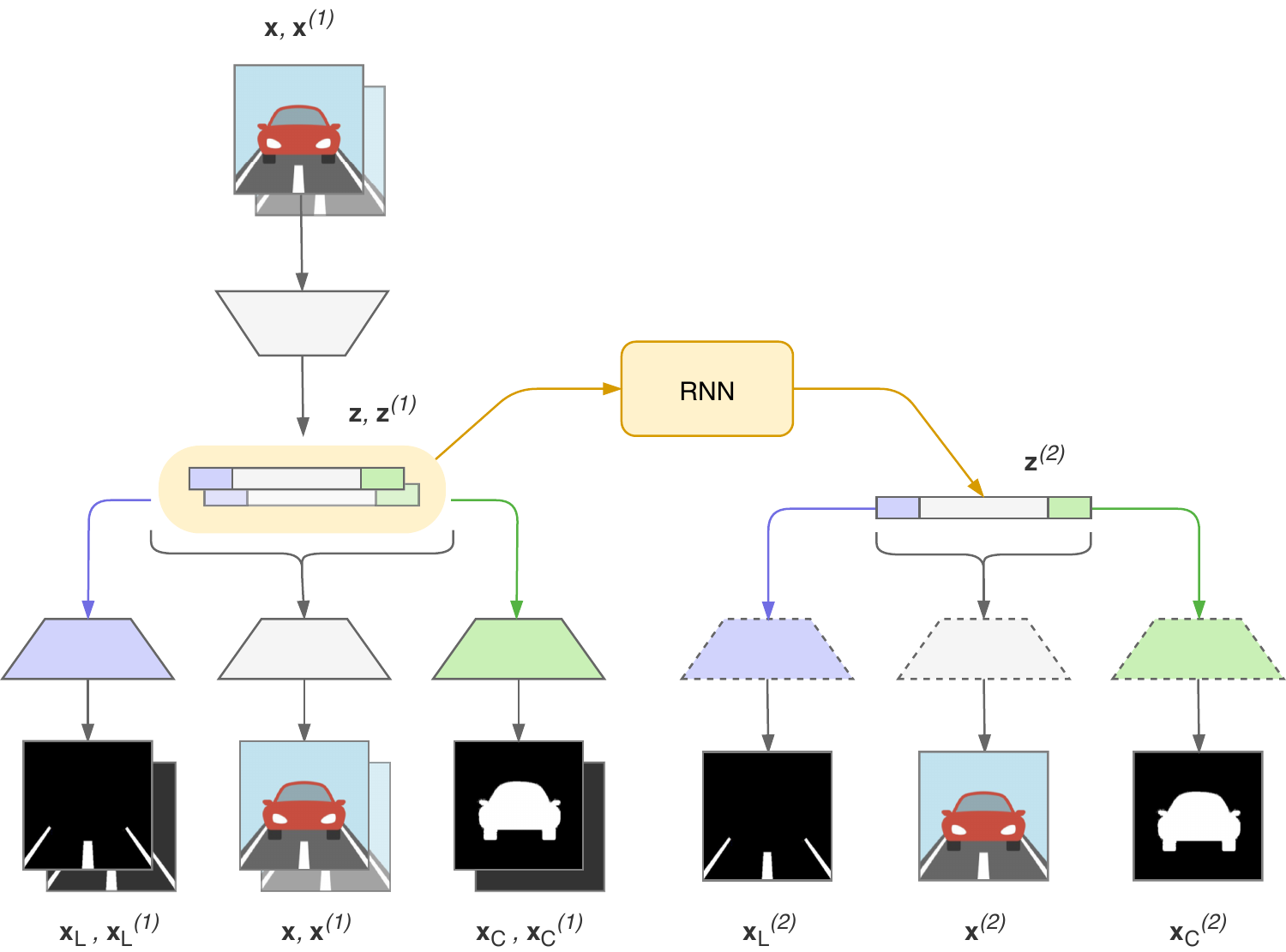}
\vspace*{5pt}
\caption{\label{f:rmvae}\small Architecture of our final temporal autoencoder (\textit{Net3}), as usual the green color denotes the \car concept and violet the \lane concept. The decoders with dashed-line border are same instances of the decoders with solid-line border.}
\end{center}
\end{figure*}
%

\subsection{Net3: Temporal Autoencoder}
\label{ss:rmvae}
\noindent
Our third model aims at including in the compact representation also the predictivity the future state of affairs. The idea here is to enforce the model to learn representations that are consistent also in the temporal dimension.
By imposing temporal consistency, we aim at further regularizing the network latent representations (whereas longer-term predictions will be the subject of the next \textit{Net4}).
Now we combine in the latent space the distinct representations of crucial concepts together with the ability to predict of how such concepts will change in future driving scenarios.

Let us introduce the notation $\vx^{(t)}$ to indicate the frame $t$ steps ahead of frame $\vx$. Similarly, $\vz^{(t)}$ refers to the the latent representation of the image $t$ steps ahead to that represented by $\vz$.
Fig. \ref{f:rmvae} shows the architecture of this temporal autoencoder (\textit{Net3}), with the final parameters described in Table \ref{t:rmvae}. The model has substantially the same architecture of the topological autoencoder (\textit{Net2}), except for an additional module based on a simple recursive neural network. The training procedure is also different: at each iteration, two subsequent frames $\vx$ and $\vx^{(1)}$ are fed as input to the common encoder, which computes two latent representations $\vz$ and $\vz^{(1)}$. These two latent vectors are fed to a RNN trained to predict the latent vector $\vz^{(2)}$ containing the representation of the consecutive frame in the sequence. All three latent vectors are then expanded using the same 3-decoders structure already seen in \textit{Net2}, so the overall model is trained to generate visual and segmented output images for all the three frames $\vx,\vx^{(1)},\vx^{(2)}$.

The novel sub-network of the model can be described by the function:
\begin{equation}
h_{\Psi}\left(\vz,\vz^{(1)}\right)\rightarrow\widetilde{\vz}\approx\vz^{(2)},
\label{e:rnn}
\end{equation}
where $h_{\Psi}$ plays the role of an autoregressive dynamic model of order 2, and it is implemented using a basic recursive neural network (RNN) \cite{E:1990} with time window of 2 and set of parameters $\Psi$. The formulation of the loss used in training the network is similar to equation \eqref{e:mloss} with additional terms for the recursive prediction:
\begin{equation}
\begin{split}
\mathcal{L}(\Theta,\Phi,\Psi|\mathcal{B})&=
\mathcal{L}(\Theta,\Phi|\mathcal{B})+\\
&\qquad
E^{\prime}_\mathrm{V}+E^{\prime}_\mathrm{C}+E^{\prime}_\mathrm{L}+\\
&\qquad
E^{\prime\prime}_\mathrm{V}+E^{\prime\prime}_\mathrm{C}+E^{\prime\prime}_\mathrm{L}
\label{e:rmloss}
\end{split}
\end{equation}
where the first term is the same loss of equation
\eqref{e:mloss}, and the expressions of the new terms are the following:
\begin{align}
E^{\prime}_\mathrm{V}=&-\lambda^{\prime}_\mathrm{V}\sum_{\vx}^\mathcal{B}\mathbb{E}_{\vz\sim{q_{\Phi}(\vz|\vx^{(1)})}}%
\left[\log{p_{\Theta_\mathrm{V}}(\vx^{(1)}|\vz)}\right],
\label{e:rmlossv}\\
\begin{split}
E^{\prime}_\mathrm{C}=&-\lambda^{\prime}_\mathrm{C}\sum_{\vx}^\mathcal{B}\mathbb{E}_{\vz_\mathrm{C}\sim\Pi_\mathrm{C}({q_{\Phi}(\vz|\vx^{(1)})})}\\
&\quad%
\left[\log{\widetilde{p}_{\Theta_\mathrm{C}}}(\vx_\mathrm{C}^{(1)}|\vz_\mathrm{C})\right],
\label{e:rmlossc}
\end{split}\\
\begin{split}
E^{\prime}_\mathrm{L}=&-\lambda^{\prime}_\mathrm{L}\sum_{\vx}^\mathcal{B}\mathbb{E}_{\vz_\mathrm{L}\sim\Pi_\mathrm{L}({q_{\Phi}(\vz|\vx^{(1)})})}\\
&\quad%
\left[\log{\widetilde{p}_{\Theta_\mathrm{L}}}(\vx_\mathrm{L}^{(1)}|\vz_\mathrm{L})\right],
\label{e:rmlossl}
\end{split}
\end{align}
\begin{align}
\begin{split}
E^{\prime\prime}_\mathrm{V}=&-\lambda^{\prime\prime}_\mathrm{V}\sum_{\vx}^\mathcal{B}\mathbb{E}_{\vz\sim{q_{\Phi}(\vz|\vx)}}\\
&\quad%
\left[\log{p_{\Theta_\mathrm{V}}\!\left(\,\vx^{(2)}\,|\,h_{\Psi}\left(\vz,g_{\Phi}(\vx^{(1)})\right)\,\right)}\right],
\end{split}
\label{e:rrmlossv}\\
\begin{split}
E^{\prime\prime}_\mathrm{C}=&-\lambda^{\prime\prime}_\mathrm{C}\sum_{\vx}^\mathcal{B}\mathbb{E}_{\vz\sim{q_{\Phi}(\vz|\vx)}}\\
&\quad%
\left[\log{\widetilde{p}_{\Theta_\mathrm{C}}}\!\left(\,\vx_\mathrm{C}^{(2)}\,|\,\Pi_\mathrm{C}\!\left(h_{\Psi}\left(\vz,g_{\Phi}(\vx^{(1)})\right)\right)\,\right)\right],
\label{e:rrmlossc}
\end{split}\\
\begin{split}
E^{\prime\prime}_\mathrm{L}=&-\lambda^{\prime\prime}_\mathrm{L}\sum_{\vx}^\mathcal{B}\mathbb{E}_{\vz\sim{q_{\Phi}(\vz|\vx)}}\\
&\quad%
\left[\log{\widetilde{p}_{\Theta_\mathrm{L}}}\!\left(\vx_\mathrm{L}^{(2)}\,|\,\Pi_\mathrm{L}\!\left(h_{\Psi}\left(\vz,g_{\Phi}(\vx^{(1)})\right)\right)\,\right)\right].
\label{e:rrmlossl}
\end{split}
\end{align}
The contributions of equations \eqref{e:rmlossv}, \eqref{e:rmlossc}, \eqref{e:rmlossl} is similar to that of $E_\mathrm{V},E_\mathrm{C},E_\mathrm{L}$, i.e. the  errors in the reconstruction of the driving scenario and conceptual entities applied to the successor of $\vz$.  The predictive relevance is carried by the errors in equations
\eqref{e:rrmlossv},
\eqref{e:rrmlossc},
\eqref{e:rrmlossl}
where the 3 decoders are applied to a latent vector resulting from the recursive sub-network $h_{\Psi}(\cdot)$.
%
%

\subsection{Net4: Recurrent Network}
\label{ss:rtime}
\noindent
%
\begin{figure*}
\begin{center}
\includegraphics[width=1.681\columnwidth]{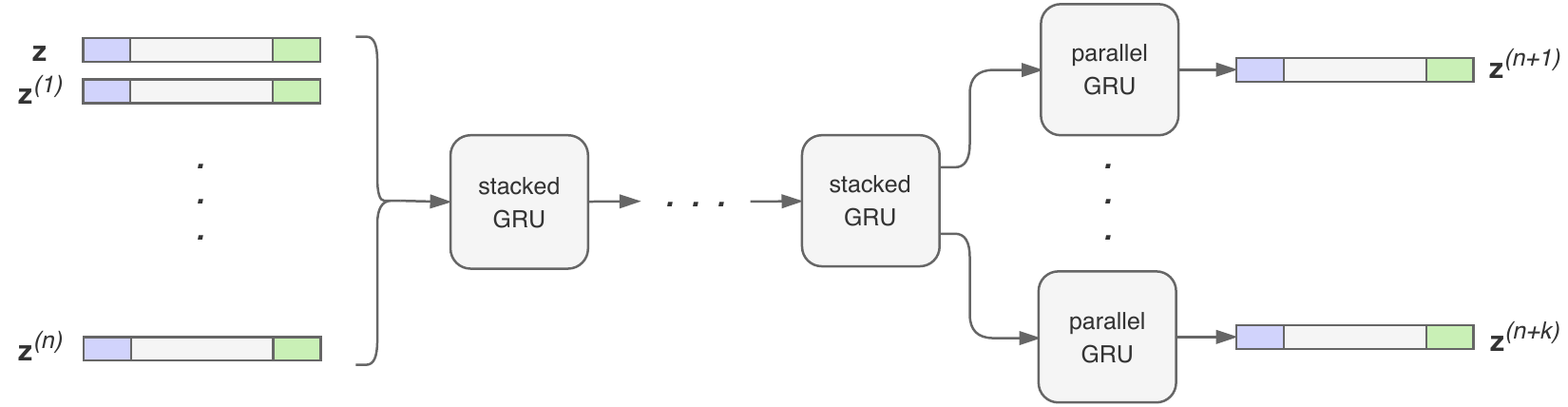}
\caption{\label{f:rtime}\small Architecture of our general recurrent model (\textit{Net4}), as usual the green color denotes the \car concept and violet the \lane concept.}
\end{center}
\end{figure*}
Once trained, the temporal autoencoder (\textit{Net3}) can be deployed in its encoding part to generate latent representations for all the images of the dataset, and the actual prediction in time can be implemented with a novel recursive network that works in the latent space only. The advantage of a very compact representation allows a much more complex recursive network than the simple $h_{\Psi}(\cdot)$ used in \textit{Net3}.

As shown in Fig. \ref{f:rtime}, the proposed recurrent network (\textit{Net4}) has a first module composed of multiple levels of stacked recurrent sub-networks, and a second module composed of multiple parallel recurrent sub-networks, each predicting future frames with increasing time steps ahead. Each stacked recurrent sub-network feeds its entire output sequence as input of the next stacked sub-network. Instead, the parallel recurrent sub-networks yield only the last output in time.
All sub-networks of this model share the same core architecture: we experimented the same simple RNN \cite{E:1990} used for $h_{\Psi}(\cdot)$, Long Short-Term Memory (LSTM) \cite{HochreiterSchmidhuber:1997} and Gated Recurrent Units (GRU) \cite{ChoEtAl:2014}, using the latter as final choice.

The mapping of the proposed recurrent model can be described as a function $r_{\Xi}:
\mathcal{Z}^{N_\mathrm{I}}
\rightarrow
\mathcal{Z}^{N_\mathrm{O}}
$
where $N_\mathrm{I}$ is the number of consecutive inputs, $N_\mathrm{O}$ is the number of predictions in time, and $\Xi$ is the set of parameters of the model. The function $r_{\Xi}(\cdot)$ is used as follows:
\begin{equation}
\begin{split}
r_{\Xi}&\left(\vz,\vz^{(1)},\cdots,\vz^{(N_\mathrm{I}-1)}\right)
\rightarrow\\
&\quad
\left[\;
\widetilde{\vz}_{1},
\widetilde{\vz}_{2},
\cdots,
\widetilde{\vz}_{N_\mathrm{O}}
\right]
\approx\\
&\quad
\left[
\vz^{(N_\mathrm{I})},
\vz^{(N_\mathrm{I}+1)},
\cdots,
\vz^{(N_\mathrm{I}+N_\mathrm{O})}
\right]
\label{e:rtime}
\end{split}
\end{equation}
In the final version of the model, we choose $N_\mathrm{I}=8$ and $N_\mathrm{O}=4$, moreover we use 2 stacked GRUs and 4 parallel GRUs, as described in Table \ref{t:rtime}.

%% file: tex/res.tex
%
%

\subsection{Dataset}
\label{ss:dset}


\noindent
We choose to train and test our models on the SYNTHIA benchmark \cite{RosEtAl:2016}. The dataset offers a large collection of image sequences representing various driving scenarios. It is realized using the game engine Unity, and it is composed of $\sim100$k frames of urban scenes recorded from a simulated camera on the windshield of the ego car. We allocated 70\% of the dataset to the training set, 25\% to validation and 5\% to the test set.

Despite being generated in 3D computer graphics, this dataset offers a wide variety of quite realistic illumination and weather conditions, resulting occasionally even in very adverse driving conditions. Each driving sequence is replicated on a set of different environmental conditions including seasons, weather and time of the day.
Moreover, the urban environment is very diverse as well, ranging from driving on freeways, through tunnels, congestion, ``NewYork-like cities'' and ``European towns'' -- as the creators of the dataset describe it.

\subsection{Results of Topological and Temporal Autoencoders}
\label{ss:r_rmvae}
\noindent

\begin{table}[t]
\scriptsize
\begin{center}
\begin{tabular}{r|c|c|c|c}
\toprule
& \multicolumn{2}{c|}{\textbf{Topological AE}} & \multicolumn{2}{c}{\textbf{Temporal AE}} \\
& \multicolumn{2}{c|}{\textbf{(\textit{Net2})}} & \multicolumn{2}{c}{\textbf{(\textit{Net3})}} \\
& \multicolumn{2}{c|}{\vspace*{-4pt}} & \multicolumn{2}{c}{} \\
& IoU car & IoU lane & IoU car & IoU lane \\
\midrule
City                & 0.7834    & 0.6487    & 0.8305    & 0.7155 \\
Freeway             & 0.7755    & 0.5840    & 0.7952    & 0.7490 \\
Sunny               & 0.7736    & 0.6283    & 0.8077    & 0.6970 \\
Dark                & 0.7682    & 0.6274    & 0.7943    & 0.7116 \\
All frames          & 0.7702    & 0.6277    & 0.7992    & 0.7062 \\
\bottomrule
\end{tabular}
\vspace*{5pt}
\caption{\label{t:iou_ae}\small IoU scores of \car and \lane classes obtained by our autoencoders. The scores are computed over all the SYNTHIA dataset, and are organized into 4 different categories of driving conditions.}
\end{center}
\end{table}

\begin{figure*}
\footnotesize
\begin{center}
\begin{tabular}{m{9pt}m{0.4287\columnwidth}@{\hspace{15pt}}m{0.4287\columnwidth}@{\hspace{15pt}}m{0.4287\columnwidth}@{\hspace{15pt}}m{0.4287\columnwidth}}
& \multicolumn{1}{c}{City} &
\multicolumn{1}{c}{Freeway} &
\multicolumn{1}{c}{Sunny} &
\multicolumn{1}{c}{Dark}
\vspace*{8pt} \\
\begin{turn}{90}Input\end{turn}&
\includegraphics[width=0.4287\columnwidth]{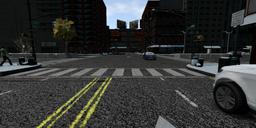} &
\includegraphics[width=0.4287\columnwidth]{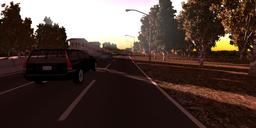} &
\includegraphics[width=0.4287\columnwidth]{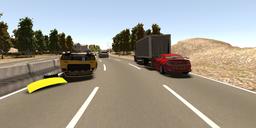} &
\includegraphics[width=0.4287\columnwidth]{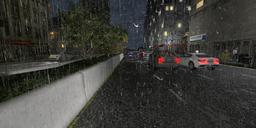} \\
\begin{turn}{90}Output\end{turn}&
\includegraphics[width=0.4287\columnwidth]{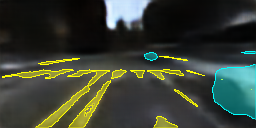} &
\includegraphics[width=0.4287\columnwidth]{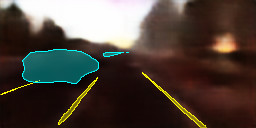} &
\includegraphics[width=0.4287\columnwidth]{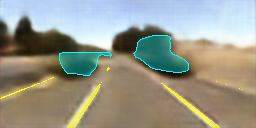} &
\includegraphics[width=0.4287\columnwidth]{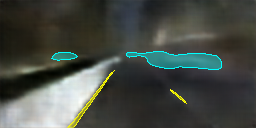} \\
\begin{turn}{90}Target\end{turn}&
\includegraphics[width=0.4287\columnwidth]{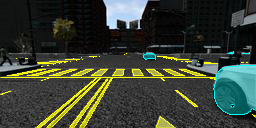} &
\includegraphics[width=0.4287\columnwidth]{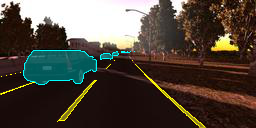} &
\includegraphics[width=0.4287\columnwidth]{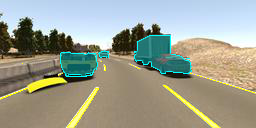} &
\includegraphics[width=0.4287\columnwidth]{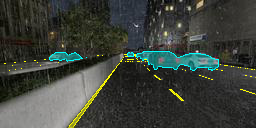} \\
\end{tabular}
\vspace*{5pt}
\caption{\label{f:pred_rmvae}\small Results of our temporal autoencoder (\textit{Net3}) in reconstructing an image. In the first row, the input frames belonging to different categories of driving scenarios. In the last row, the same input frames plotted with a colored overlay showing the target \car entities in cyan and the \lane entities in yellow. In the center row, the output of the network.}
\end{center}
\end{figure*}

\begin{figure}
\scriptsize
\begin{center}
\begin{tabular}{
m{9pt}@{\hspace{0pt}}
m{9pt}
m{0.3667\columnwidth}@{\hspace{20pt}}
m{0.3667\columnwidth}}
\begin{turn}{90}frame A\end{turn} &
\begin{turn}{90}raw\end{turn} &
\includegraphics[width=0.3667\columnwidth]{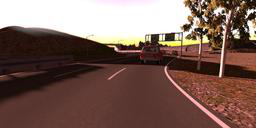} &
\includegraphics[width=0.3667\columnwidth]{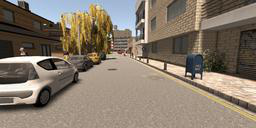} \\
\begin{turn}{90}frame A\end{turn} &
\begin{turn}{90}segmented\end{turn} &
\includegraphics[width=0.3667\columnwidth]{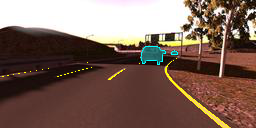} &
\includegraphics[width=0.3667\columnwidth]{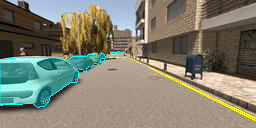} \\
&&
\includegraphics[width=0.3667\columnwidth]{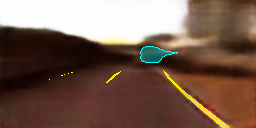} &
\includegraphics[width=0.3667\columnwidth]{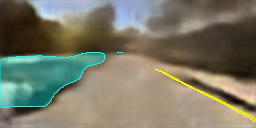} \\
&&
\includegraphics[width=0.3667\columnwidth]{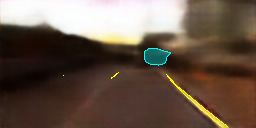} &
\includegraphics[width=0.3667\columnwidth]{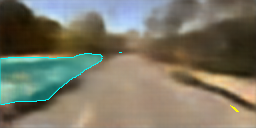} \\
&&
\includegraphics[width=0.3667\columnwidth]{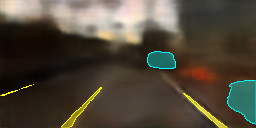} &
\includegraphics[width=0.3667\columnwidth]{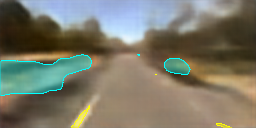} \\
&&
\includegraphics[width=0.3667\columnwidth]{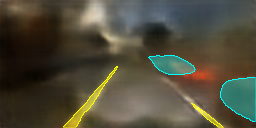} &
\includegraphics[width=0.3667\columnwidth]{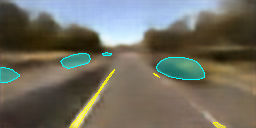} \\
&&
\includegraphics[width=0.3667\columnwidth]{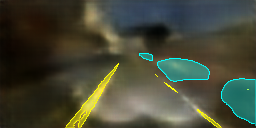} &
\includegraphics[width=0.3667\columnwidth]{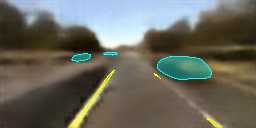} \\
\begin{turn}{90}frame B\end{turn} &
\begin{turn}{90}segmented\end{turn} &
\includegraphics[width=0.3667\columnwidth]{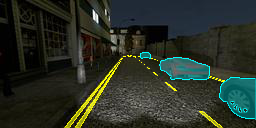} &
\includegraphics[width=0.3667\columnwidth]{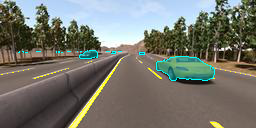} \\
\begin{turn}{90}frame B\end{turn} &
\begin{turn}{90}raw\end{turn} &
\includegraphics[width=0.3667\columnwidth]{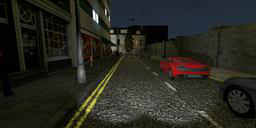} &
\includegraphics[width=0.3667\columnwidth]{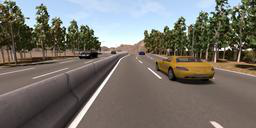} \\
\end{tabular}
\caption{\label{f:interp_rmvae}\small Examples of interpolation between latent representations learned by our temporal autoencoder (\textit{Net3}), for two different driving scenarios. The first 2 rows show the first input frame, with and without the colored overlay highlighting the \car and \lane entities. The last 2 rows show the second input frame. The 5 central rows are the result of the linear interpolation between the latent representations of the inputs.}
\end{center}
\end{figure}

\noindent
Here we present the comparison between the results of the topological and the temporal autoencoder (\textit{Net2} and \textit{Net3}), the final networks were trained for 200 epochs.
Table \ref{t:iou_ae} displays the IoU \textit{(Intersection over Union)} scores for the \car and \lane classes, for each model. The scores are grouped into 4 categories: driving in urban environments, driving on freeways, driving in sunny conditions, driving with adverse lighting conditions or during bad weather. The Table includes also the general scores for all the frames in the dataset.
These scores nicely show how the temporal model (\textit{Net3}) is able to learn a more consistent latent representation, with respect to the topological model (\textit{Net2}), in all the categories of driving sequences. However, it is also evident how the task of recognizing the \car concept always ends up in better scores compared to the \lane concept. An explanation of why the latter task is more difficult can be the very low ratio of pixels belonging to the class of \lane over the entire image size, and consequently how easily the lane markings get occluded by other elements in the scene.


We would like to stress that the purpose of our networks is not mere segmentation of visual input. The segmentation operation is to be considered as a support task, used to enforce the networks to learn a more robust latent space representation, which now is explicitly taking into consideration two of the concepts that are fundamental to the driving tasks.

To visually appreciate the representation learned by the temporal autoencoder (\textit{Net3}), Fig. \ref{f:pred_rmvae} shows the images produced by the model for 4 different input images, one for each category of driving sequences mentioned above. We take an input image (showed on the left of the Figure) and make the network produce its corresponding latent representation. The latent vector is passed to the decoders of the network to reconstruct the initial image (output showed on the center of the Figure), in which the class of \car concepts is colored in cyan and the class of \lane concepts in yellow. The images on the right are just displayed as a reference, they show the target images with the colored overlay of the two classes.

To further visualize the performance of our temporal autoencoder (\textit{Net3}), we present the result of interpolating between different latent spaces. In Fig. \ref{f:interp_rmvae}, each column shows what happens when taking the latent representation of a first frame (first row in the Figure) and linearly interpolate it with the latent representation of a second frame (last row). We generate 5 intermediate latent vectors, which are passed to the decoders of the temporal autoencoder to produce novel frames. The images are a smooth and gradual shift from the first input to the second, and successfully provide new plausible driving scenarios never seen before by the network.

\subsection{Results of Recurrent Network}
\label{ss:r_rnn}
\noindent

\begin{table*}
\scriptsize
\begin{center}
\begin{tabular}{r|c|c|c|c|c|c|c|c}
\toprule &
\multicolumn{2}{c|}{\textbf{Frame 9}} & 
\multicolumn{2}{c|}{\textbf{Frame 10}} & 
\multicolumn{2}{c|}{\textbf{Frame 11}} & 
\multicolumn{2}{c}{\textbf{Frame 12}} \\
& IoU car & IoU lane & IoU car & IoU lane
& IoU car & IoU lane & IoU car & IoU lane \\
\midrule
City                & 0.7543 & 0.5692 & 0.7173 & 0.5472 & 0.6799 & 0.5421 & 0.6381 & 0.5220 \\
Freeway             & 0.6928 & 0.5197 & 0.6336 & 0.4698 & 0.5967 & 0.4487 & 0.5589 & 0.4296 \\
Sunny               & 0.7223 & 0.5338 & 0.6768 & 0.5001 & 0.6661 & 0.4831 & 0.6106 & 0.4693 \\
Dark                & 0.7000 & 0.5226 & 0.6570 & 0.5120 & 0.6130 & 0.5014 & 0.5834 & 0.4832 \\
All frames          & 0.7078 & 0.5268 & 0.6639 & 0.5075 & 0.6315 & 0.4946 & 0.5931 & 0.4782 \\
\bottomrule
\end{tabular}
\vspace*{5pt}
\caption{\label{t:iou_rtime}\small IoU scores of \car and \lane classes obtained by our recurrent network (\textit{Net4}) when predicting 4 future frames from an input sequence of 8 frames. The scores are computed over all the SYNTHIA dataset, and are organized into 4 different categories of driving conditions.}
\end{center}
\end{table*}

\begin{figure*}
\footnotesize
\begin{center}
\begin{tabular}{
m{9pt}
m{9pt}
m{0.4110\columnwidth}@{\hspace{5pt}}
m{0.4110\columnwidth}@{\hspace{5pt}}
m{0.4110\columnwidth}@{\hspace{5pt}}
m{0.4110\columnwidth}}
& & \multicolumn{1}{c}{Frame 9} &
\multicolumn{1}{c}{Frame 10} &
\multicolumn{1}{c}{Frame 11} &
\multicolumn{1}{c}{Frame 12}
\vspace*{8pt} \\
\multirow{5}{*}{\begin{turn}{90}City\end{turn}}&
\begin{turn}{90}Output\end{turn}&
\includegraphics[width=0.4110\columnwidth]{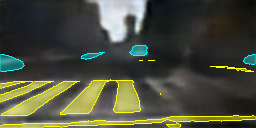} &
\includegraphics[width=0.4110\columnwidth]{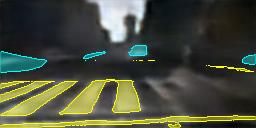} &
\includegraphics[width=0.4110\columnwidth]{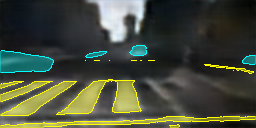} &
\includegraphics[width=0.4110\columnwidth]{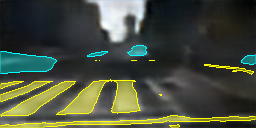} \\
& \begin{turn}{90}Target\end{turn}&
\includegraphics[width=0.4110\columnwidth]{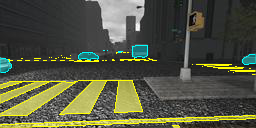} &
\includegraphics[width=0.4110\columnwidth]{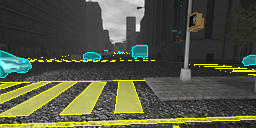} &
\includegraphics[width=0.4110\columnwidth]{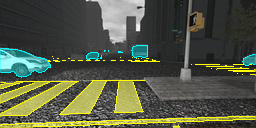} &
\includegraphics[width=0.4110\columnwidth]{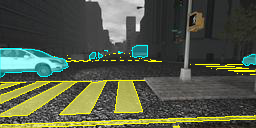} \\\\
\multirow{5}{*}{\begin{turn}{90}Freeway\end{turn}}&
\begin{turn}{90}Output\end{turn}&
\includegraphics[width=0.4110\columnwidth]{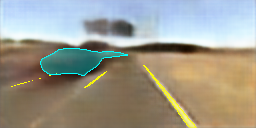} &
\includegraphics[width=0.4110\columnwidth]{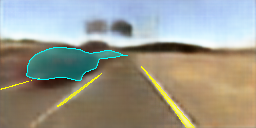} &
\includegraphics[width=0.4110\columnwidth]{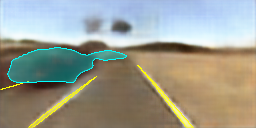} &
\includegraphics[width=0.4110\columnwidth]{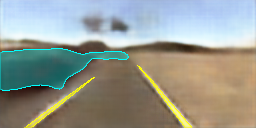} \\
& \begin{turn}{90}Target\end{turn}&
\includegraphics[width=0.4110\columnwidth]{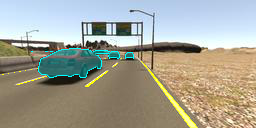} &
\includegraphics[width=0.4110\columnwidth]{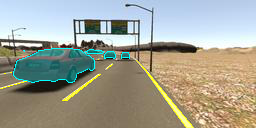} &
\includegraphics[width=0.4110\columnwidth]{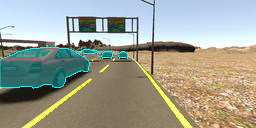} &
\includegraphics[width=0.4110\columnwidth]{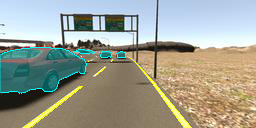} \\\\
\multirow{4}{*}{\begin{turn}{90}Sunny\end{turn}}&
\begin{turn}{90}Output\end{turn}&
\includegraphics[width=0.4110\columnwidth]{imgs/pred/S06WR_000321_pred_8.png} &
\includegraphics[width=0.4110\columnwidth]{imgs/pred/S06WR_000321_pred_9.png} &
\includegraphics[width=0.4110\columnwidth]{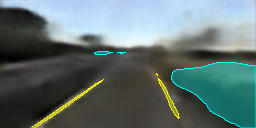} &
\includegraphics[width=0.4110\columnwidth]{imgs/pred/S06WR_000321_pred_11.png} \\
& \begin{turn}{90}Target\end{turn}&
\includegraphics[width=0.4110\columnwidth]{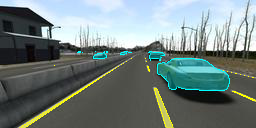} &
\includegraphics[width=0.4110\columnwidth]{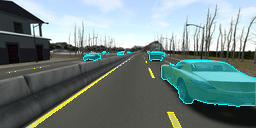} &
\includegraphics[width=0.4110\columnwidth]{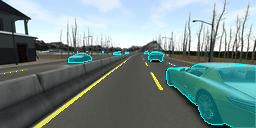} &
\includegraphics[width=0.4110\columnwidth]{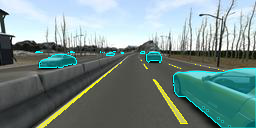} \\\\
\multirow{4}{*}{\begin{turn}{90}Dark\end{turn}}&
\begin{turn}{90}Output\end{turn}&
\includegraphics[width=0.4110\columnwidth]{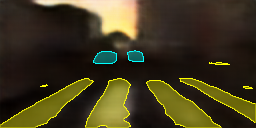} &
\includegraphics[width=0.4110\columnwidth]{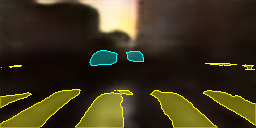} &
\includegraphics[width=0.4110\columnwidth]{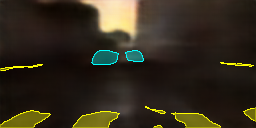} &
\includegraphics[width=0.4110\columnwidth]{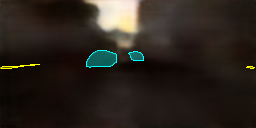} \\
& \begin{turn}{90}Target\end{turn}&
\includegraphics[width=0.4110\columnwidth]{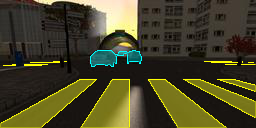} &
\includegraphics[width=0.4110\columnwidth]{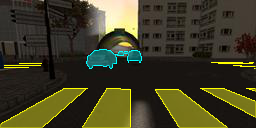} &
\includegraphics[width=0.4110\columnwidth]{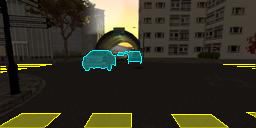} &
\includegraphics[width=0.4110\columnwidth]{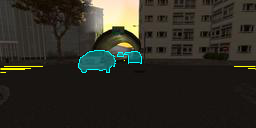} \\

\end{tabular}
\vspace*{5pt}
\caption{\label{f:pred_rtime}\small Results of our recurrent model (\textit{Net4}) in predicting 4 future frames from an input sequence of 8 frames. Odd rows show the output of the network, even rows the corresponding target frames.}
\end{center}
\end{figure*}

\begin{figure*}[t]
\scriptsize
\begin{center}
\begin{tabular}{cc@{\hspace{30pt}}cc}
Output & Target & Output & Target \vspace*{8pt} \\
\includegraphics[width=0.3589\columnwidth]{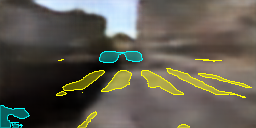} &
\includegraphics[width=0.3589\columnwidth]{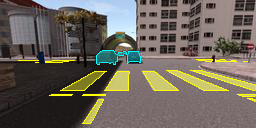} &
\includegraphics[width=0.3589\columnwidth]{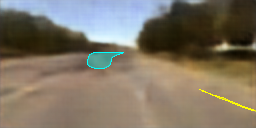} &
\includegraphics[width=0.3589\columnwidth]{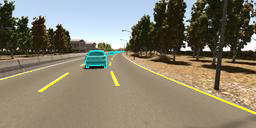} \\
\includegraphics[width=0.3589\columnwidth]{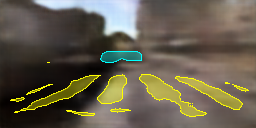} &
\includegraphics[width=0.3589\columnwidth]{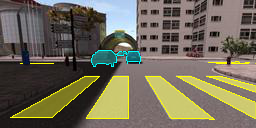} &
\includegraphics[width=0.3589\columnwidth]{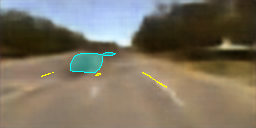} &
\includegraphics[width=0.3589\columnwidth]{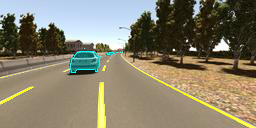} \\
\includegraphics[width=0.3589\columnwidth]{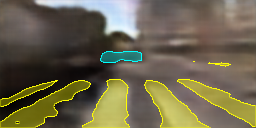} &
\includegraphics[width=0.3589\columnwidth]{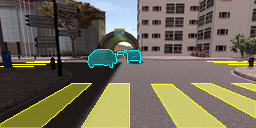} &
\includegraphics[width=0.3589\columnwidth]{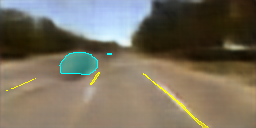} &
\includegraphics[width=0.3589\columnwidth]{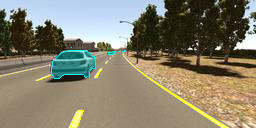} \\
\includegraphics[width=0.3589\columnwidth]{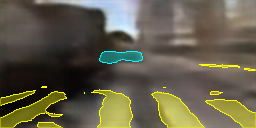} &
\includegraphics[width=0.3589\columnwidth]{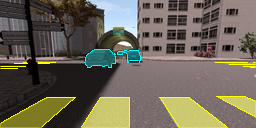} &
\includegraphics[width=0.3589\columnwidth]{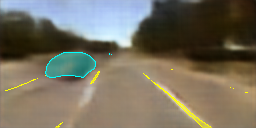} &
\includegraphics[width=0.3589\columnwidth]{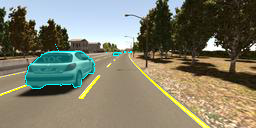} \\
\includegraphics[width=0.3589\columnwidth]{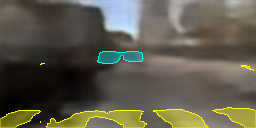} &
\includegraphics[width=0.3589\columnwidth]{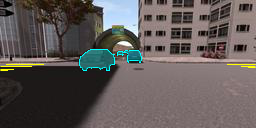} &
\includegraphics[width=0.3589\columnwidth]{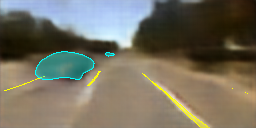} &
\includegraphics[width=0.3589\columnwidth]{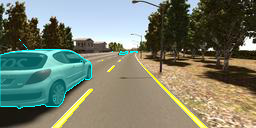} \\
\includegraphics[width=0.3589\columnwidth]{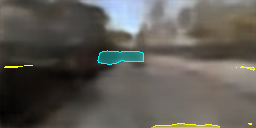} &
\includegraphics[width=0.3589\columnwidth]{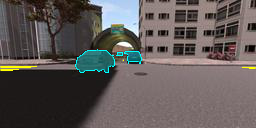} &
\includegraphics[width=0.3589\columnwidth]{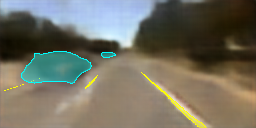} &
\includegraphics[width=0.3589\columnwidth]{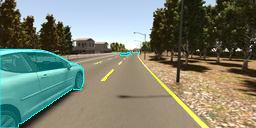} \\
\includegraphics[width=0.3589\columnwidth]{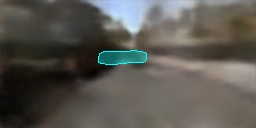} &
\includegraphics[width=0.3589\columnwidth]{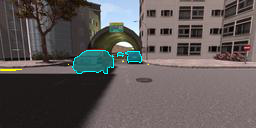} &
\includegraphics[width=0.3589\columnwidth]{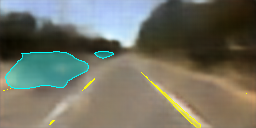} &
\includegraphics[width=0.3589\columnwidth]{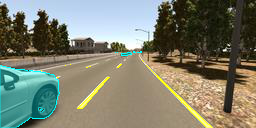} \\
\includegraphics[width=0.3589\columnwidth]{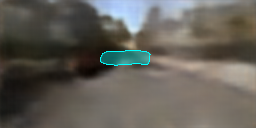} &
\includegraphics[width=0.3589\columnwidth]{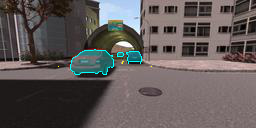} &
\includegraphics[width=0.3589\columnwidth]{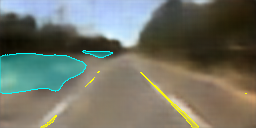} &
\includegraphics[width=0.3589\columnwidth]{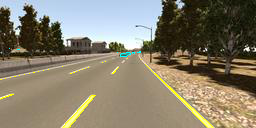} \\
\includegraphics[width=0.3589\columnwidth]{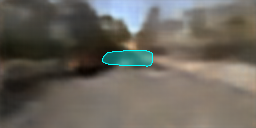} &
\includegraphics[width=0.3589\columnwidth]{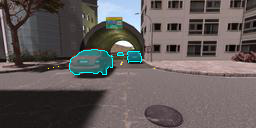} &
\includegraphics[width=0.3589\columnwidth]{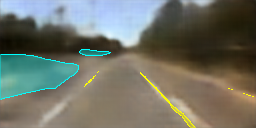} &
\includegraphics[width=0.3589\columnwidth]{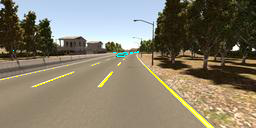} \\
\end{tabular}
\vspace*{5pt}
\caption{\label{f:halluc}\small Examples of mental imagery on our recurrent model (\textit{Net4}), for two different driving scenarios. Odd columns show the result of iteratively feed the predicted output of the model back as input of the next iteration. Even columns are a reference on the corresponding frames.}
\end{center}
\end{figure*}

\noindent
Lastly, we present the results of our final network, the recursive model (\textit{Net4}), trained for 100 epochs on a dataset of latent representations computed by the temporal autoencoder (\textit{Net3}) over the SYNTHIA frames.

Table \ref{t:iou_rtime} contains the IoU scores obtained by the model in the different categories of driving sequences used before. As described in \S\ref{ss:r_rnn}, the network takes as input a sequence of 8 frames and predicts the 4 subsequent frames. The Table shows the scores for the 4 predicted frames, separated as usual in the \car and \lane classes.
It is immediate to note the \car scores are always higher than the \lane scores, just like we saw in Table \ref{t:iou_ae}. However, the \car predictions worsen more significantly for the distant frames, where we can see a decay of 16\%, while the \lane scores lose only 9\%. This result can be explained by the fact that, generally, in a driving sequence the lane markings change in a more smooth and predictable way with respect to the cars, which can for example suddenly modify their trajectory.

Fig. \ref{f:pred_rtime} depicts the visual results of predictions, one for each category of driving sequences. We include in the Figure the 4 predicted frames and their corresponding target frames, we omit to show the 8 input frames in order to keep the Figure easy to read. The model is able to predict an overtake maneuver from the left as well as from the right (``freeway'' and ``sunny'' cases). Another interesting result is the different kind of predictions when facing a crosswalk, in the ``city'' scenario there is a car moving perpendicularly to the lane of the ego car, so the network correctly predicts to hold still at the cross walk. In the ``dark'' scenario there are cars driving in the same direction of the ego car, therefore the model predicts to not stop at the crosswalk move forward.

As a further test, we tried to replicate the phenomenon of mental imagery using our recurrent model (\textit{Net4}), i.e. the network is called iteratively and at each iteration the output is fed back as input of the next iteration. In our specific case we choose to take the 1st of the 4 output vectors and use it as the 8th input vector of the next iteration. Fig. \ref{f:halluc} presents the results of 9 iterations of imagery for two different scenarios, along with the corresponding reference frames (the input images are, again, omitted for practical reasons). Note that, while the imagery process must inevitably start with all input frames taken from the dataset, the results provided in the Figure are obtained from forward iterations, that is when all input vectors are computed by the network as results of previous iterations.
In both driving scenarios, it is possible to appreciate how the model is able to predict a quite plausible future from just its own representation of the world.

\subsection{Latent Representations}
\label{ss:lat}
\noindent

\begin{table}[t]
\scriptsize
\begin{center}
\begin{tabular}{r|c|c}
\toprule
& Temporal      & Predictivity \\
& coherence     & error \\
\midrule
Variational AE (\textit{Net1})     & 0.299   & 0.186 \\
Topological AE (\textit{Net2})     & 0.297   & 0.189 \\
Temporal AE (\textit{Net3})        & 0.180   & 0.077 \\
\bottomrule
\end{tabular}
\vspace*{5pt}
\caption{\label{t:lat_eval}\small Simple statistics on the latent representations learned by our 3 autoencoder models. For both indicators, the lower the better.}
\end{center}
\end{table}

\begin{figure*}[t]
\footnotesize
\begin{center}
\begin{tabular}{c@{\hspace{10pt}}c@{\hspace{10pt}}c@{\hspace{10pt}}c}
Frame & Car & Other visual features & Lane \\\\
\includegraphics[height=0.2112\columnwidth]{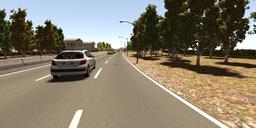} &
\includegraphics[height=0.2232\columnwidth]{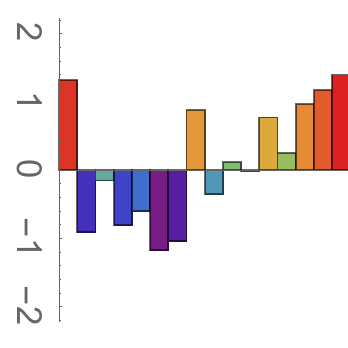} &
\includegraphics[height=0.2112\columnwidth]{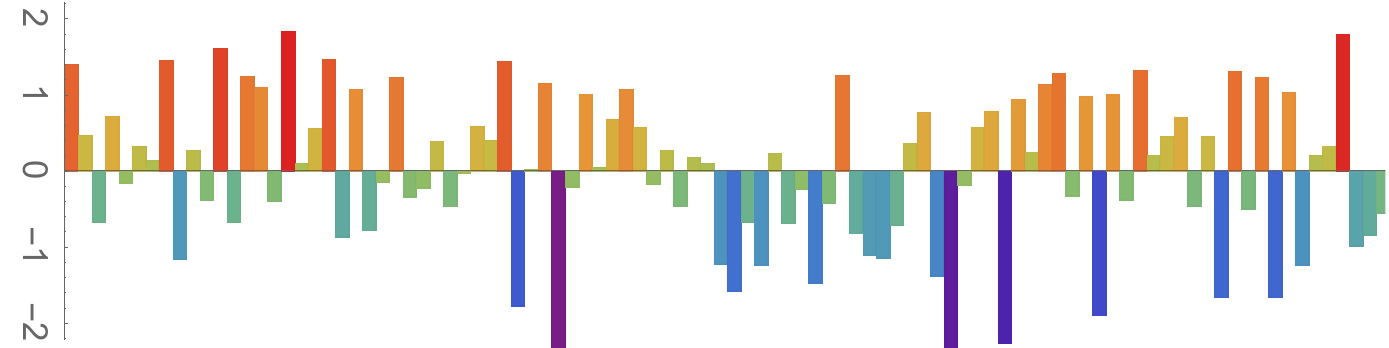} &
\includegraphics[height=0.2232\columnwidth]{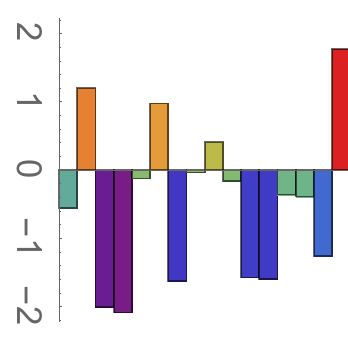} \\
\includegraphics[height=0.2112\columnwidth]{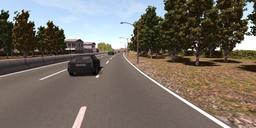} &
\includegraphics[height=0.2232\columnwidth]{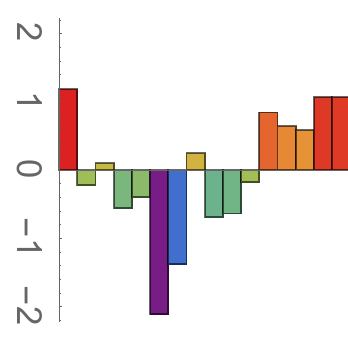} &
\includegraphics[height=0.2112\columnwidth]{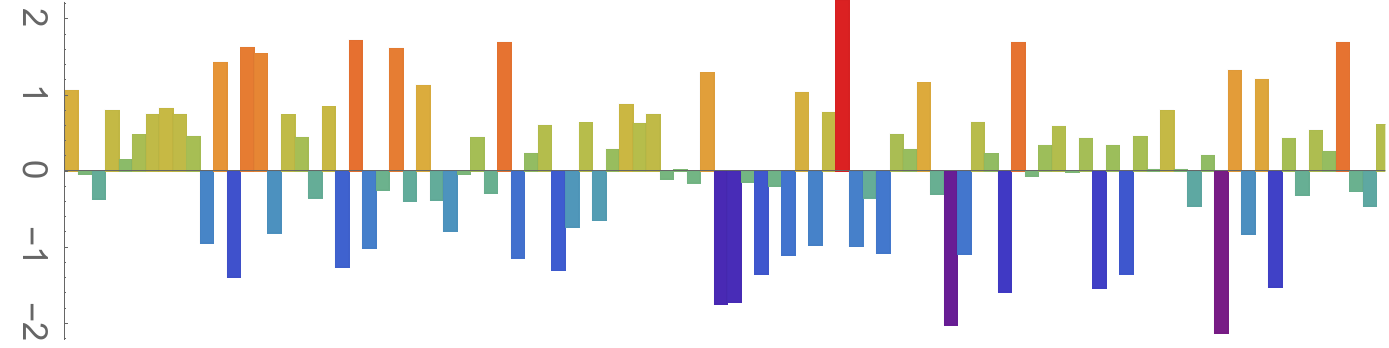} &
\includegraphics[height=0.2232\columnwidth]{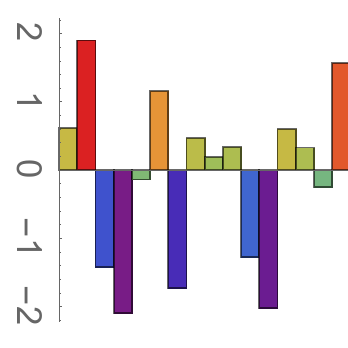} \\
\includegraphics[height=0.2112\columnwidth]{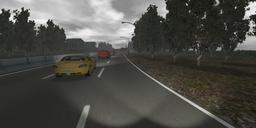} &
\includegraphics[height=0.2232\columnwidth]{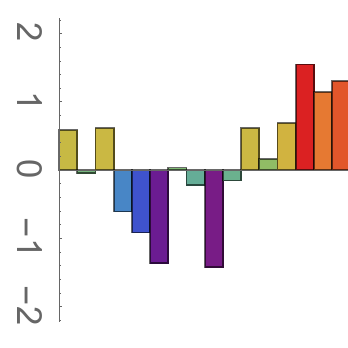} &
\includegraphics[height=0.2112\columnwidth]{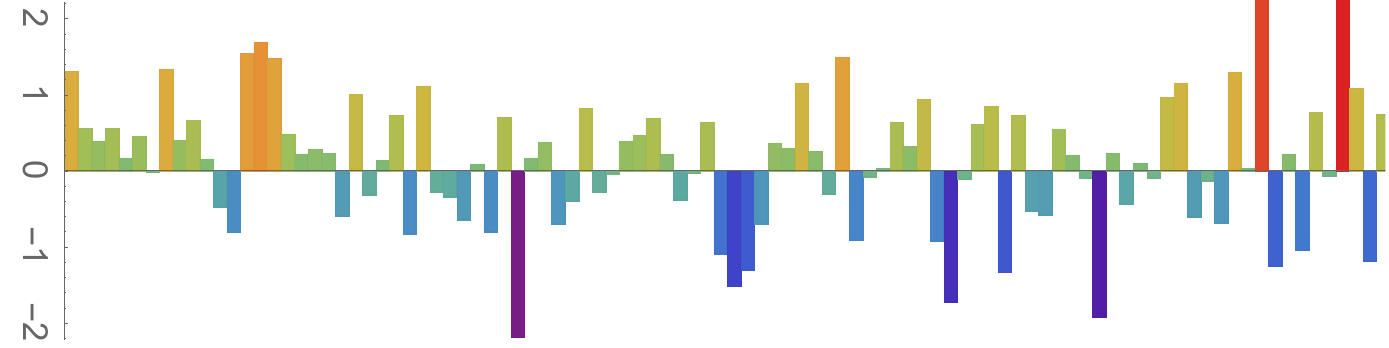} &
\includegraphics[height=0.2232\columnwidth]{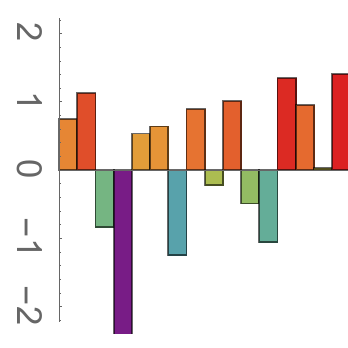} \\
\includegraphics[height=0.2112\columnwidth]{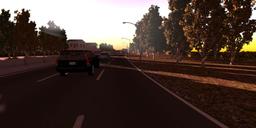} &
\includegraphics[height=0.2232\columnwidth]{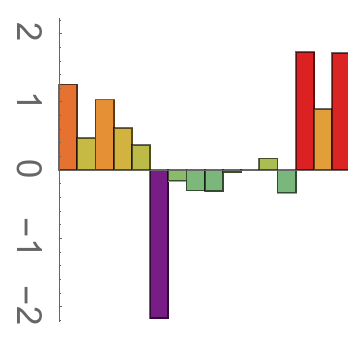} &
\includegraphics[height=0.2112\columnwidth]{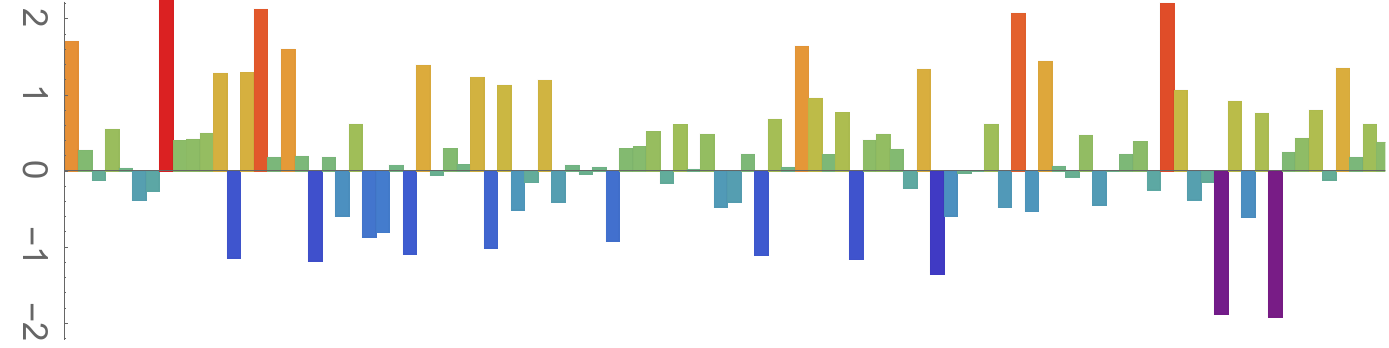} &
\includegraphics[height=0.2232\columnwidth]{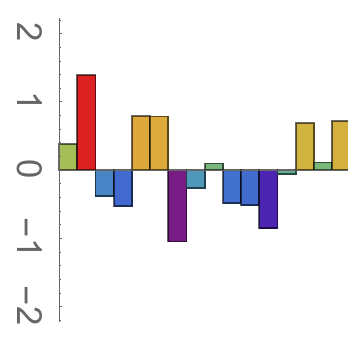} \\
\end{tabular}
\vspace*{5pt}
\caption{\label{f:lat_mvae}\small Visualization of the latent representation learned by our topological autoencoder (\textit{Net2}). Each row shows the values of the 128 neurons of the latent representation of the image on the left. The neurons corresponding to the \car and \lane concepts are plotted separately.}
\end{center}
\end{figure*}

\begin{figure*}[t]
\footnotesize
\begin{center}
\begin{tabular}{c@{\hspace{10pt}}c@{\hspace{10pt}}c@{\hspace{10pt}}c}
Frame & Car & Other visual features & Lane \\\\
\includegraphics[height=0.2112\columnwidth]{imgs/latent/1_S06SM_000338_RGB_FL.jpg} &
\includegraphics[height=0.2232\columnwidth]{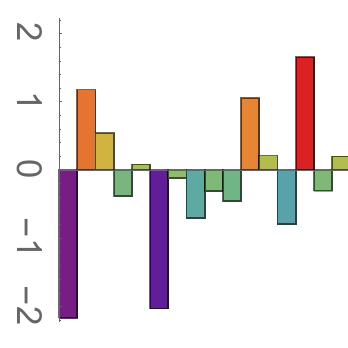} &
\includegraphics[height=0.2112\columnwidth]{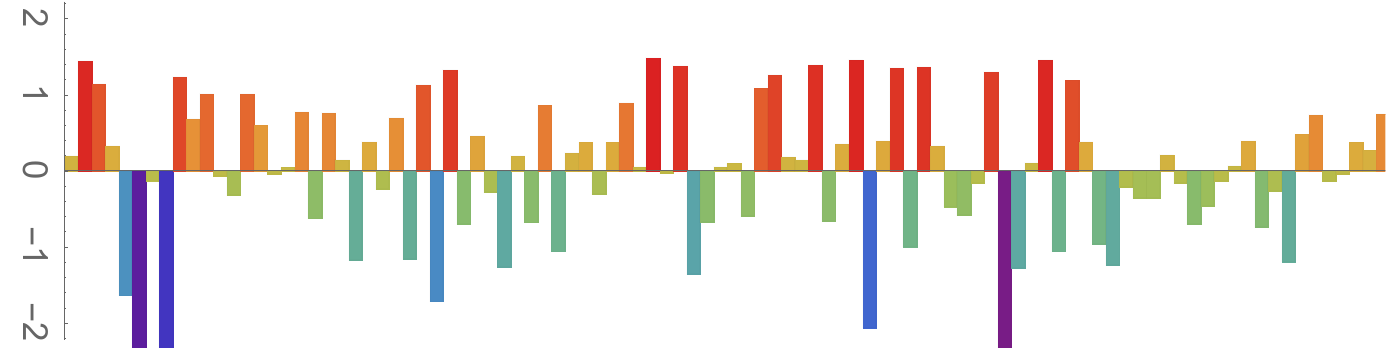} &
\includegraphics[height=0.2232\columnwidth]{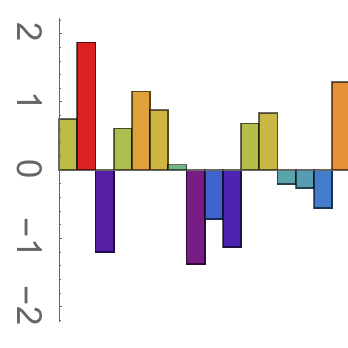} \\
\includegraphics[height=0.2112\columnwidth]{imgs/latent/2_S06SP_000345_RGB_FL.jpg} &
\includegraphics[height=0.2232\columnwidth]{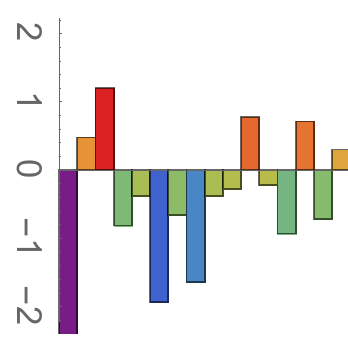} &
\includegraphics[height=0.2112\columnwidth]{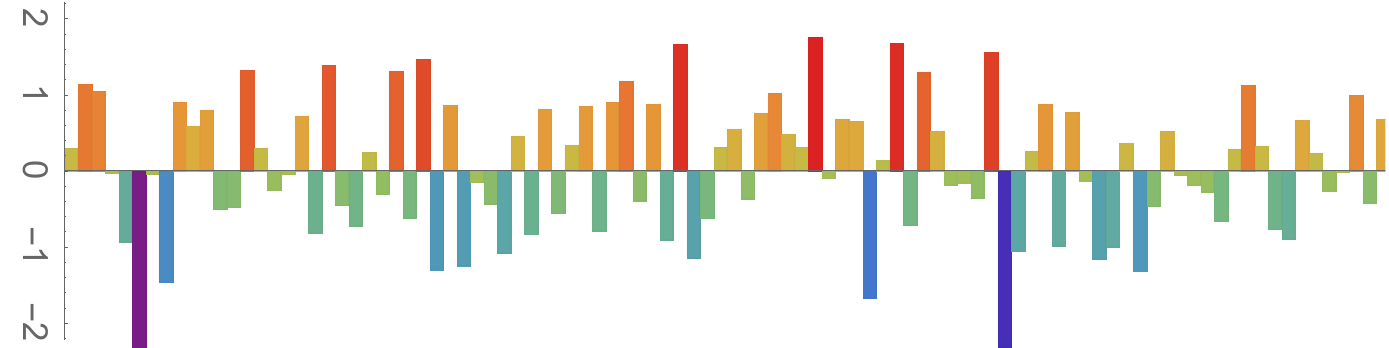} &
\includegraphics[height=0.2232\columnwidth]{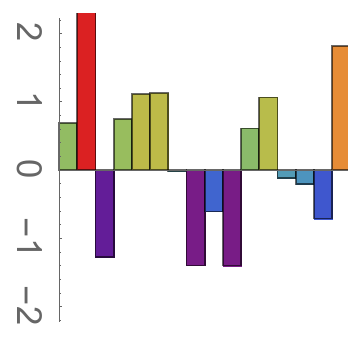} \\
\includegraphics[height=0.2112\columnwidth]{imgs/latent/3_S06FG_000243_RGB_FL.jpg} &
\includegraphics[height=0.2232\columnwidth]{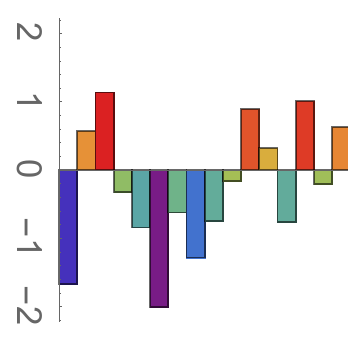} &
\includegraphics[height=0.2112\columnwidth]{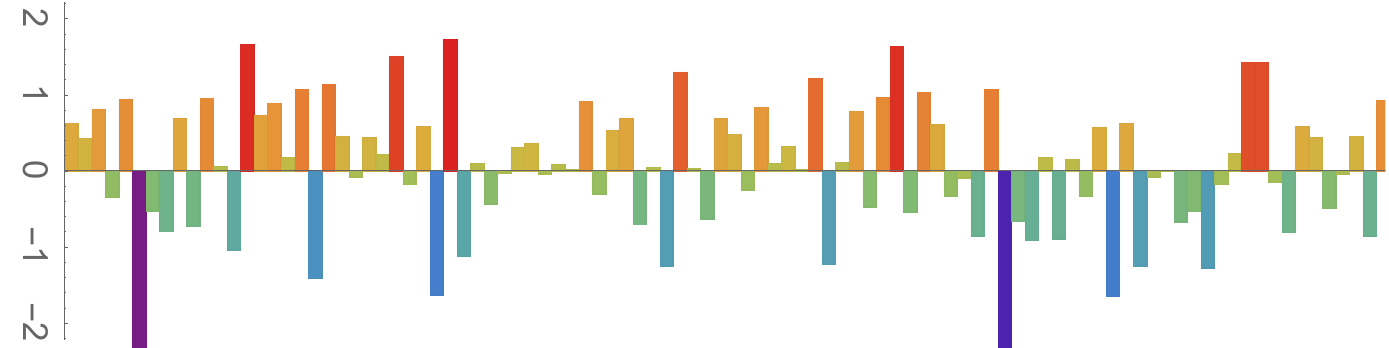} &
\includegraphics[height=0.2232\columnwidth]{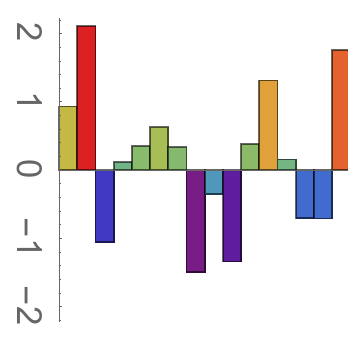} \\
\includegraphics[height=0.2112\columnwidth]{imgs/latent/4_S06DW_000316_RGB_FL.jpg} &
\includegraphics[height=0.2232\columnwidth]{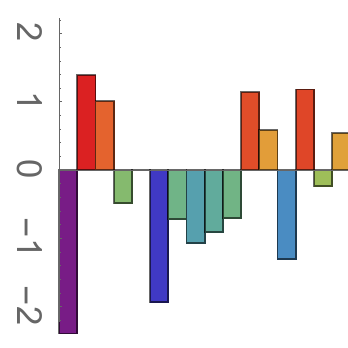} &
\includegraphics[height=0.2112\columnwidth]{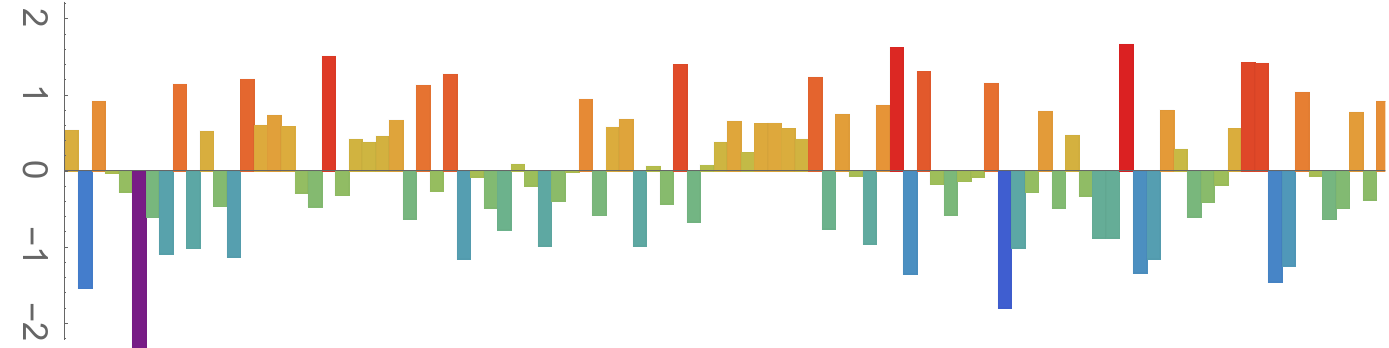} &
\includegraphics[height=0.2232\columnwidth]{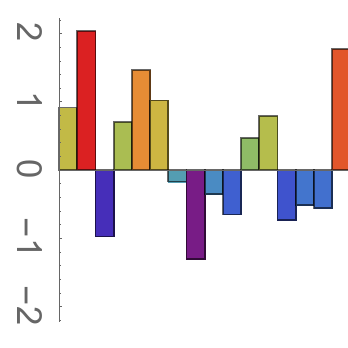} \\
\end{tabular}
\vspace*{5pt}
\caption{\label{f:lat_rmvae}\small Visualization of the latent representation learned by our temporal autoencoder (\textit{Net3}). Each row shows the values of the 128 neurons of the latent representation of the image on the left. The neurons corresponding to the \car and \lane concepts are plotted separately.}
\end{center}
\end{figure*}

\noindent
We performed additional validations of the latent representations learned by the models with simple statistical evaluations of their ability to predict in the future, and their consistency with respect to the temporal dynamics. This sort of evaluation has been useful in the development of the models, in Table \ref{t:lat_eval} we report the results obtained with the 3 final encoder--decoder models. The first indicator $\xi$ evaluates the degree of temporal coherence by the ratio between the difference of two latent vectors that are contiguous in time, and the variance over the entire dataset of latent $\mathcal{Z}$. The evaluation is done independently for each component of the latent vector, and then averaged:
\begin{equation}
\xi_\mathcal{Z}=\frac{1}{N_\mathrm{V}}\sum_i^{N_\mathrm{V}}\frac{%
\sum_{\vz\in\mathcal{Z}}\left(z_i-\zeta^1(z_i)\right)^2}{M\upsilon^{(i)}_\mathcal{Z}},
\label{e:xi}
\end{equation}
where $z_i$ is the $i$-th element of $\vz$, $\upsilon^{(i)}_\mathcal{Z}$ is the $i$-th element of the variance vector of $\vz$ over $\mathcal{Z}$, and $M$ is the cardinality of $\mathcal{Z}$. The ideal temporal coherence should be $\xi_\mathcal{Z}=0$. 

The second indicator $\rho$ is the mean square of the residual obtained when using two consecutive latent vectors to predict one neuron of a third vector, by linear regression. In order to make computation time acceptable, this index is computed on a subspace $\mathcal{Z}^{\prime}$ ten times smaller than $\mathcal{Z}$.  By calling $\varepsilon(\vec{A},\vec{b})$ the residual of the least squares approximation of the normal equation $\vec{A}\vec{x}=\vec{b}$, $\rho$ can be written as follows:
\begin{equation}
\rho_{\mathcal{Z}^{\prime}}=\frac{1}{N_\mathrm{V}}\sum_i^{N_\mathrm{V}}\varepsilon\left(%
\begin{bmatrix}
\cdots & \cdots\\
\vz & \zeta^1(\vz)\\
\cdots & \cdots\\
\end{bmatrix}
_{\vz\in{\mathcal{Z}^{\prime}}}
,
\begin{bmatrix}
\cdots\\
\zeta^2(z_i)\\
\cdots\\
\end{bmatrix}
_{\vz\in{\mathcal{Z}^{\prime}}}
\right)
\label{e:rho}
\end{equation}

As a final visualization of the performance of our models, we present an inspection of the latent representations learned by the topological and temporal autoencoders (\textit{Net2} and \textit{Net3}), Fig. \ref{f:lat_mvae} and \ref{f:lat_rmvae}  respectively. In each Figure, the first row shows 4 images depicting the same driving scenario under different lighting conditions. For each input image, we plot the values of the 128 neurons composing the latent encoding computed by the model, separating the 16 neurons representing the \car entities (second row of the Figure), the 16 neurons representing the \lane entities (last row) and the remaining 96 neurons representing generic visual features (third row). Ideally, only the generic 96 neurons should change in the 4 cases, because the input images differ only in the lighting conditions while having the same \car and \lane entities. Practically, the cars and lane markings are not exactly the same in all the 4 input images, therefore a margin of error in the latent representations is considered acceptable.

Comparing Fig. \ref{f:lat_mvae} and \ref{f:lat_rmvae} it is immediately clear how the temporal autoencoder (\textit{Net3}) learns a  more robust representation, the variation in the neurons encoding the \car and \lane concepts is minimum. Also the variation in the general 96 neurons are very localized, the neurons exhibit a similar overall distribution, and this fits with the fact that the 4 frames have the same surrounding (the trees, the soil on the right).
Conversely, the representation learned by topological autoencoder (\textit{Net2}) does not appear as consistent. The \car and \lane neurons change significantly for each input frame, and even the other 96 visual features do not share any particular pattern in the 4 cases.

%% file: tex/end.tex
\noindent
This paper presented an algorithm for perception of driving scenarios, our work takes inspiration from some principles on how the brain of the current best drivers -- humans -- works. We did not have to invent anything from scratch: the deep learning framework already offers tools that -- when used in a proper way -- can implement the neurocognitive theories we aim at.  Specifically, we used autoencoders for implementing the theoretical idea of coding perceptual concepts using the lowest possible dimension, as in neural convergence-divergence zones. Then we followed the theory of predictive brain by encouraging the probabilistic representation learned by the autoencoder to capture information about the future.

Compared to other research on perception of driving scenarios based on autoencoder, our approach is unique in combining the conventional training of the encoder by minimizing the loss on the decoded reconstruction of the input, with the ability of the encoded representation to perform prediction in time. Experimenting with the SYNTHIA dataset, we were able to converge the visual input down to a representation of just 16 neurons for the concept of \car and other 16 neurons for the \lane concept. With this compact representation we achieved good performances in predicting future frames up to 4 time steps ahead.

The system described in this paper is certainly not a complete solution to perceptual understanding of driving scenarios, it is a strategy. In several driving contexts there are more concepts than \car and \lane, such as pedestrians and cyclists, not taken into account in our system. An even more crucial task is the projection of the inner representation to more complex spaces than the visual and conceptual ones: the space of actions. In fact, the ongoing research in our group is in decoding the latent representation into the two-dimensional space of affordable longitudinal and lateral controls.

%% file: tex/ack.tex
\noindent 
This work was developed inside the EU Horizon 2020 Dreams4Cars Research and Innovation Action, supported by the European Commission under Grant 731593.
The Authors want also to thank the Deep Learning Lab at the ProM Facility in Rovereto, Italy, for supporting this research with computational resources funded by Fondazione CARITRO.

%% file: tex/appx.tex
\noindent
The variational inference framework takes up the issue of approximating the probability distribution $p(\vx)$ of a high dimensional random variable $\vx\in\mathcal{X}$. This approximation can be performed by a neural network such as that in equation \eqref{e:decoder}.  The neural network by itself is deterministic, but its output distribution can be easily computed as follows:
\begin{equation}
p_{\Theta}(\vx|\vz)=\mathcal{N}\left(\vx|f_{\Theta}(\vz),\vec\sigma^2
\mathbf{I}\right),\label{e:pxz}
\end{equation}
where $\mathcal{N}(\vx|\vec\mu,\vec\sigma)$ is the Gaussian function in $\vx$, with mean $\vec\mu$ and standard deviation $\vec\sigma$.  Using this last equation it is now possible to express the desired approximation of $p(\vx)$:
\begin{equation}
p_{\Theta}(\vx)=\int{p_{\Theta}(\vx,\vz)d\vz}=
\int{p_{\Theta}(\vx|\vz)p(\vz)d\vz}.
\label{e:px}
\end{equation}
It is immediate to recognize that the kind of neural network performing the function $f_\Theta(\cdot)$ is exactly the decoder part in the autoencoder, corresponding to the divergence zone in the CDZ neurocognitive concept.  In the case when $\mathcal{X}$ is the domain of images, $f_\Theta(\cdot)$ comprises a first layer that rearranges the low-dimension variable $\vx$ in a two dimensional geometry, followed by a stack of deconvolutions, up to the final geometry of the $\vx$ images.

In equation \eqref{e:px} there is clearly no clue on what the distribution $p(\vz)$ might be, but the idea behind variational autoencoder is to introduce an auxiliary distribution $q$ from which to sample $\vz$, and it is made by an additional neural network. Ideally, this network should provide the posterior probability $p_\Theta(\vz|\vx)$ -- which is unknown -- and should be a network like the kind of equation \eqref{e:encoder}. Its probability distribution is:
\begin{equation}
q_{\Phi}(\vz|\vx)=\mathcal{N}\left(\vz|g_{\Phi}(\vx),\vec\sigma^2
\mathbf{I}\right).\label{e:pzx}
\end{equation}
While the network $f_\Theta(\cdot)$ behaves as decoder, the network $g_{\Phi}(\cdot)$ corresponds to the encoder part in the autoencoder, projecting the high-dimensional variable $\vx$ into the low dimensional space $\mathcal{Z}$. It continues to play the role of the convergence zone in the CDZ idea.

The measure of how well $p_{\Theta}(\vx)$ approximates $p(\vx)$ for a set of $\vx_i\in\mathcal{D}$ sampled in a dataset $\mathcal{D}$ is given by the log-likelihood:
\begin{equation}
\ell(\Theta|\mathcal{D})=%
\sum_{\vx_i\in\mathcal{D}}\log\int{p_{\Theta}(\vx_i|\vz)p(\vz)d\vz}.
\label{e:likelihood}
\end{equation}
This equation cannot be solved because of the unknown $p(\vz)$, and here comes the help of the auxiliary probability $q_{\Phi}(\vz|\vx)$. Each term of the summation in equation \eqref{e:likelihood} can be rewritten as follows:
\begin{align}
\ell(\Theta|\vx)&=\log\int{p_{\Theta}(\vx,\vz)d\vz}\nonumber\\
&=\log\int\frac{p_{\Theta}(\vx,\vz)q_{\Phi}(\vz|\vx)}{q_{\Phi}(\vz|\vx)}d\vz\nonumber\\
&=\log\mathbb{E}_{\vz\sim{q_{\Phi}(\vz|\vx)}}\left[\frac{p_{\Theta}(\vx,\vz)}{q_{\Phi}(\vz|\vx)}\right],
\label{e:likelihoodq}
\end{align}
where in the last passage we used the expectation operator $\mathbb{E}[\cdot]$.  Being the $\log$ function concave, we can now apply Jensen's inequality:
\begin{equation}
\begin{split}
\ell(\Theta,\Phi|\vx)%
&=\log\mathbb{E}_{\vz\sim{q_{\Phi}(\vz|\vx)}}\left[\frac{p_{\Theta}(\vx,\vz)}{q_{\Phi}(\vz|\vx)}\right]
\\
&\ge\mathbb{E}_{\vz\sim{q_{\Phi}(\vz|\vx)}}\left[\log{p_{\Theta}(\vx,\vz)}\right]-\\
&\qquad
\mathbb{E}_{\vz\sim{q_{\Phi}(\vz|\vx)}}\left[\log{q_{\Phi}(\vz|\vx)}\right].
\end{split}
\label{e:elbo}
\end{equation}
Since the derivation in the last equation is smaller or at least equal to $\ell(\Theta|\vx)$, it is called the \textit{variational lower bound}, or \textit{evidence lower bound} (ELBO). Note that now in $\ell(\Theta,\Phi|\vx)$ there is also the dependency from the parameters $\Phi$ of the second neural network defined in \eqref{e:pzx}.

It is possible to rearrange further $\ell(\Theta,\Phi|\vx)$ in order to have $p_{\Theta}(\vx|\vz)$ instead of $p_{\Theta}(\vx,\vz)$ in equation \eqref{e:elbo}, moreover, we can now introduce the \textit{loss function} $\mathcal{L}(\Theta,\Phi|\vx)$ as the value to be minimized in order to maximize
ELBO:
\begin{equation}
\begin{split}
\mathcal{L}(\Theta,\Phi|\vx)&=-\ell(\Theta,\Phi|\vx)\\
&=
-\int{q_{\Phi}(\vz|\vx)\log\frac{p_{\Theta}(\vx,\vz)}{q_{\Phi}(\vz|\vx)}d\vz}\\
&=
-\int{q_{\Phi}(\vz|\vx)\log\frac{p_{\Theta}(\vx|\vz)p_{\Theta}(\vz)}{q_{\Phi}(\vz|\vx)}d\vz}\\
&=
\Delta_{\mathrm{KL}}\big(q_{\Phi}(\vz|\vx)\|p_{\Theta}(\vz)\big)
-\\
&\qquad
\mathbb{E}_{\vz\sim{q_{\Phi}(\vz|\vx)}}\left[\log{p_{\Theta}(\vx|\vz)}\right],\label{e:loss}
\end{split}
\end{equation}
where the last step uses the  Kullback-Leibler divergence $\Delta_{\mathrm{KL}}$.
Still, this formulation seems to be intractable because it contains the term $p_{\Theta}(\vz)$, but there is a simple analytical formulation of the Kullback-Leibler divergence in the Gaussian case (see Appendix B in \cite{KingmaWelling:2014}):
\begin{equation}
\begin{split}
\Delta_{\mathrm{KL}}&\Big(q_{\Phi}(\vz|\vx)\|p(\vz)\Big)=\\
&\quad
-\frac{1}{2}\sum_{i=1}^{Z}\left(1+\log\big(\sigma_i^2)\big)%
-\mu_j^2-\sigma_i^2\right),
\label{e:klN}
\end{split}
\end{equation}
where $\mu_i$ and $\sigma_i$ are the $i$-th components of the mean and variance
of $\vz$ given by $q_{\Phi}(\vz|\vx)$.

%% file: tex/table.tex
\begin{table}[H]
\scriptsize
\begin{center}
\begin{tabular}{m{80pt}m{60pt}m{50pt}}
\toprule
Encoder &       convolution     & $7\times7\times16$ \\
&               convolution     & $7\times7\times32$ \\
&               convolution     & $5\times5\times32$ \\
&               convolution     & $5\times5\times32$ \\
&               dense           & 2048 \\
&               dense           & 512 \\
\midrule
Latent space &                        & 128 \\
\midrule
Decoder &      dense           & 2048 \\
& dense           & 4096 \\
& deconvolution   & $5\times5\times32$ \\
& deconvolution   & $5\times5\times32$ \\
& deconvolution   & $7\times7\times16$ \\
& deconvolution   & $7\times7\times3$ \\
\midrule
Total parameters    & & 18 million \\
\bottomrule
\end{tabular}
\vspace*{5pt}
\caption{\label{t:vae}\footnotesize Parameters describing the architecture of the variational autoencoder (\textit{Net1}).}
\end{center}
\end{table}


\begin{table}[H]
\scriptsize
\begin{center}
\begin{tabular}{m{80pt}m{60pt}m{50pt}}
\toprule
Encoder &       convolution     & $7\times7\times16$ \\
&               convolution     & $7\times7\times32$ \\
&               convolution     & $5\times5\times32$ \\
&               convolution     & $5\times5\times32$ \\
&               dense           & 2048 \\
&               dense           & 512 \\
\midrule
Latent space &                        & $[ 16, 96, 16 ]$ \\
\midrule
Each decoder &      dense           & 2048 \\
& dense           & 4096 \\
& deconvolution   & $5\times5\times32$ \\
& deconvolution   & $5\times5\times32$ \\
& deconvolution   & $7\times7\times16$ \\
& deconvolution   & $7\times7\times3$ \\
\midrule
Total parameters    & & 35 million \\
\bottomrule\end{tabular}
\vspace*{5pt}
\caption{\label{t:mvae}\footnotesize Parameters describing the architecture of the topological autoencoder (\textit{Net2}).}
\end{center}
\end{table}


\begin{table}[H]
\scriptsize
\begin{center}
\begin{tabular}{m{80pt}m{60pt}m{50pt}}
\toprule
Encoder &       convolution     & $7\times7\times16$ \\
&               convolution     & $7\times7\times32$ \\
&               convolution     & $5\times5\times32$ \\
&               convolution     & $5\times5\times32$ \\
&               dense           & 2048 \\
&               dense           & 512 \\
\midrule
Latent space &                        & $[ 16, 96, 16 ]$ \\
\midrule
Recurrent layer &                     & $128\times2\rightarrow128$ \\
\midrule
Each of the 3 &      dense           & 2048 \\
individual decoders & dense           & 4096 \\
& deconvolution   & $5\times5\times32$ \\
& deconvolution   & $5\times5\times32$ \\
& deconvolution   & $7\times7\times16$ \\
& deconvolution   & $7\times7\times3$ \\
\midrule
Total parameters    & & 35 million \\
\bottomrule\end{tabular}
\vspace*{5pt}
\caption{\label{t:rmvae}\footnotesize Parameters describing the architecture of the temporal autoencoder (\textit{Net3}).}
\end{center}
\end{table}


\begin{table}[H]
\scriptsize
\begin{center}
\begin{tabular}{m{80pt}m{40pt}m{70pt}}
\toprule
Stacked recurrency  &   GRU     & $128\times8\rightarrow128\times8$ \\
                    &   GRU     & $128\times8\rightarrow128\times8$ \\
\midrule
Parallel recurrency &   GRU     & $128\times8\rightarrow128$ \\
                    &   GRU     & $128\times8\rightarrow128$ \\
                    &   GRU     & $128\times8\rightarrow128$ \\
                    &   GRU     & $128\times8\rightarrow128$ \\
\midrule
Total parameters    &           &  600.000 \\
\bottomrule\end{tabular}
\vspace*{5pt}
\caption{\label{t:rtime}\footnotesize Parameters describing the architecture of the recurrent network (\textit{Net4}).}
\end{center}
\end{table}